\documentclass{article} % For LaTeX2e
\usepackage{iclr2020_conference,times}

\usepackage{amsmath,amsfonts,bm}

\def\eqref#1{equation~\ref{#1}}
\def\1{\bm{1}}

\DeclareMathAlphabet{\mathsfit}{\encodingdefault}{\sfdefault}{m}{sl}
\SetMathAlphabet{\mathsfit}{bold}{\encodingdefault}{\sfdefault}{bx}{n}

\usepackage{hyperref}
\usepackage{url}

\usepackage[utf8]{inputenc} % allow utf-8 input
\usepackage[T1]{fontenc}    % use 8-bit T1 fonts
\usepackage{hyperref}       % hyperlinks
\usepackage{url}            % simple URL typesetting
\usepackage{booktabs}       % professional-quality tables
\usepackage{amsfonts}       % blackboard math symbols
\usepackage{nicefrac}       % compact symbols for 1/2, etc.
\usepackage{microtype}      % microtypography

\usepackage{graphicx}
\usepackage{subcaption}
\usepackage{amsmath}
\usepackage{algorithm}
\usepackage{algorithmic}
\usepackage{amssymb}
\usepackage{array}
\usepackage{multirow}
\usepackage{enumitem}

\newcommand{\vocab}[1]{\textit{\textbf{#1}}}

\title{Measuring the Reliability of \\ Reinforcement Learning Algorithms}

\author{
Stephanie C.Y. Chan,\textsuperscript{1}\thanks{Work done as part of the Google AI Residency} \hspace{1mm}
 Samuel Fishman,\textsuperscript{1} 
 John Canny,\textsuperscript{1, 2}
 Anoop Korattikara,\textsuperscript{1} \\
 \textbf{\& Sergio Guadarrama\textsuperscript{1}}
\\
\textsuperscript{1}{Google Research} 
\textsuperscript{2}{Berkeley EECS}\\ 
\texttt{\{scychan,sfishman,canny,kbanoop,sguada\}@google.com}
}

\iclrfinalcopy % Uncomment for camera-ready version, but NOT for submission.
\begin{document}

\maketitle

\begin{abstract}
    Lack of reliability is a well-known issue for reinforcement learning (RL) algorithms. This problem has gained increasing attention in recent years, and efforts to improve it have grown substantially. To aid RL researchers and production users with the evaluation and improvement of reliability, we propose a set of metrics that quantitatively measure different aspects of reliability. In this work, we focus on variability and risk, both during training and after learning (on a fixed policy). We designed these metrics to be general-purpose, and we also designed complementary statistical tests to enable rigorous comparisons on these metrics. In this paper, we first describe the desired properties of the metrics and their design, the aspects of reliability that they measure, and their applicability to different scenarios. We then describe the statistical tests and make additional practical recommendations for reporting results. The metrics and accompanying statistical tools have been made available as an open-source library.\footnote{https://github.com/google-research/rl-reliability-metrics} We apply our metrics to a set of common RL algorithms and environments, compare them, and analyze the results.
\end{abstract}

\section{Introduction}

Reinforcement learning (RL) algorithms, especially Deep RL algorithms, tend to be highly variable in performance and considerably sensitive to a range of different factors, including implementation details, hyper-parameters, choice of environments, and even random seeds \citep{henderson_deep_2017}. This variability hinders reproducible research, and can be costly or even dangerous for real-world applications. Furthermore, it impedes scientific progress in the field when practitioners cannot reliably evaluate or predict the performance of any particular algorithm, compare different algorithms, or even compare different implementations of the same algorithm.

Recently, \citet{henderson_deep_2017} has performed a detailed analysis of reliability for several policy gradient algorithms, while \citet{duan_benchmarking_2016} has benchmarked average performance of different continuous-control algorithms. In other related work, \citet{colas_how_2018} have provided a detailed analysis on power analyses for mean performance in RL, and \citet{colas_hitchhikers_2019} provide a comprehensive primer on statistical testing for mean and median performance in RL.

In this work, we aim to devise a set of metrics that measure reliability of RL algorithms. Our analysis distinguishes between several typical modes to evaluate RL performance: "evaluation during training", which is computed over the course of training, vs. "evaluation after learning", which is evaluated on a fixed policy after it has been trained. These metrics are also designed to measure different aspects of reliability, e.g. reproducibility (variability \textit{across} training runs and variability \textit{across} rollouts of a fixed policy) or stability (variability \textit{within} training runs). Additionally, the metrics capture multiple aspects of variability -- dispersion (the width of a distribution), and risk (the heaviness and extremity of the lower tail of a distribution).

Standardized measures of reliability can benefit the field of RL by allowing RL practitioners to compare algorithms in a rigorous and consistent way. This in turn allows the field to measure progress, and also informs the selection of algorithms for both research and production environments. By measuring various aspects of reliability, we can also identify particular strengths and weaknesses of algorithms, allowing users to pinpoint specific areas of improvement. 

In this paper, in addition to describing these reliability metrics, we also present practical recommendations for statistical tests to compare metric results and how to report the results more generally. As examples, we apply these metrics to a set of algorithms and environments (discrete and continuous, off-policy and on-policy). We have released the code used in this paper as an open-source Python package to ease the adoption of these metrics and their complementary statistics.

\vspace{5mm}

\renewcommand\labelitemi{\tiny$\bullet$}
\newcommand{\pboxone}{\parbox[t]{0.06\hsize}}
\newcommand{\pboxtwo}{\parbox[t]{0.2\hsize}}
\newcommand{\pboxthree}{\parbox[t]{0.3\hsize}}

\begin{table*}[ht]
\centering
\begin{tabular}[ht]{|c|c|c|c|}
    \hline 
    & & \textbf{Dispersion (D)} & \textbf{Risk (R)} \\
    \hline
     \pboxone{\multirow{2}{*}[-0.4ex]{\rotatebox[origin=c]{90}{\parbox{2.0cm}{\textsc{During training}}}}}
        & \pboxtwo{\centering \smallskip \textbf{Across Time (T) (within training runs)}}
                        & \pboxthree{\centering \smallskip IQR$^*$ within windows, after detrending}
                        & \pboxthree{\centering \smallskip \textbf{Short-term:} CVaR$^\dagger$ on first-order differences\smallbreak \textbf{Long-term}: CVaR$^\dagger$ on Drawdown\medbreak}\\
        \cline{2-4}
        & \pboxtwo{\centering \smallskip \textbf{Across Runs (R)}}
                            & \pboxthree{\centering \smallskip IQR$^*$ across training runs, after low-pass filtering. \smallskip}
                            & \pboxthree{\centering \smallskip CVaR$^\dagger$ across runs}\\
    \hline
    \pboxone{\multirow{1}{*}{\rotatebox[origin=c]{90}{\parbox{1.7cm}{\textsc{After learning}}}}} &
    \pboxtwo{\centering \smallskip \textbf{Across rollouts on a Fixed Policy (F) \smallskip}}
                        & \pboxthree{\centering \smallskip IQR$^*$ across rollouts for a fixed policy}
                        & \pboxthree{\centering \smallskip CVaR$^\dagger$ across rollouts for a fixed policy\medbreak\medbreak\medbreak\medbreak}\\
    \hline
\end{tabular}
\caption{Summary of our proposed reliability metrics. For evaluation \textsc{during training}, which measures reliability over the course of training an algorithm, the inputs to the metrics are the performance curves of an algorithm, evaluated at regular intervals during a single training run (or on a set of training runs). For evaluation \textsc{after learning}, which measures reliability of an already-trained policy, the inputs to the metrics are the performance scores of a set of rollouts of that fixed policy. 
$^*$IQR: inter-quartile range. $^\dagger$CVaR: conditional value at risk.}
\label{table_of_metrics}
\end{table*}

\section{Reliability Metrics}

We target three different axes of variability, and two different measures of variability along each axis. We denote each of these by a letter, and each metric as a combination of an axis + a measure, e.g. "DR" for "Dispersion Across Runs". See Table \ref{table_of_metrics} for a summary. Please see Appendix \ref{assumptions_and_definitions} for more detailed definitions of the terms used here.

\subsection{Axes of variability}
\label{Axes of variability}

Our metrics target the following three axes of variability. The first two capture reliability "during training", while the last captures reliability of a fixed policy "after learning".

\paragraph{During training: Across Time (T)}

In the setting of evaluation during training, one desirable property for an RL algorithm is to be stable "across time" within each training run. In general, smooth monotonic improvement is preferable to noisy fluctuations around a positive trend, or unpredictable swings in performance.

This type of stability is important for several reasons. During learning, especially when deployed for real applications, it can be costly or even dangerous for an algorithm to have unpredictable levels of performance. Even in cases where bouts of poor performance do not directly cause harm, e.g. if training in simulation, high instability implies that algorithms have to be check-pointed and evaluated more frequently in order to catch the peak performance of the algorithm, which can be expensive. Furthermore, while training, it can be a waste of computational resources to train an unstable algorithm that tends to forget previously learned behaviors.

\paragraph{During training: Across Runs (R)} 

During training, RL algorithms should have easily and consistently reproducible performances across multiple training runs. Depending on the components that we allow to vary across training runs, this variability can encapsulate the algorithm's sensitivity to a variety of factors, such as: random seed and initialization of the optimization, random seed and initialization of the environment, implementation details, and hyper-parameter settings. Depending on the goals of the analysis, these factors can be held constant or allowed to vary, in order to disentangle the contribution of each factor to variability in training performance. High variability on any of these dimensions leads to unpredictable performance, and also requires a large search in order to find a model with good performance.

% For example, if we keep the implementation and hyper-parameters constant across different instances of training, then this variability is a measure the algorithm's sensitivity to random seeds. 

% If we instead vary the hyper-parameter settings across the training runs, then this variability across training runs measures reproducibility across hyper-parameter settings. In many cases, for practical reasons, stochasticity from the training procedure (i.e. the optimization) is not easily disentangled from stochasticity from the environment.

\paragraph{After learning: Across rollouts of a fixed policy (F)}

When evaluating a fixed policy, a natural concern is the variability in performance across multiple rollouts of that fixed policy. Each rollout may be specified e.g. in terms of a number of actions, environment steps, or episodes. Generally, this metric measures sensitivity to both stochasticity from the environment and stochasticity from the training procedure (the optimization). Practitioners may sometimes wish to keep one or the other constant if it is important to disentangle the two factors (e.g. holding constant the random seed of the environment while allowing the random seed controlling optimization to vary across rollouts).

\subsection{Measures of variability}

For each axis of variability, we have two kinds of measures: dispersion and risk.

\paragraph{Dispersion}

Dispersion is the width of the distribution. To measure dispersion, we use "robust statistics" such as the \vocab{Inter-quartile range (IQR)} (i.e. the difference between the 75th and 25th percentiles) and the \vocab{Median absolute deviation from the median (MAD)}, which are more robust statistics and don't require assuming normality of the distributions. \footnote{Note that our aim here is to measure the variability of the distribution, rather than to characterize the uncertainty in estimating a statistical parameter of that distribution. Therefore, confidence intervals and other similar methods are not suitable for the aim of measuring dispersion.} We prefer to use IQR over MAD, because it is more appropriate for asymmetric distributions \citep{peter_j._rousseeuw_alternatives_1993}.

\paragraph{Risk}

In many cases, we are concerned about the worst-case scenarios. Therefore, we define risk as the heaviness and extent of the lower tail of the distribution. This is complementary to measures of dispersion like IQR, which cuts off the tails of the distribution.
% For example, during learning, we would like to ensure that the agent is constantly improving as it interacts with the world. Or, for example, in the case of hyper-parameter optimization, we would like to know as early as possible whether a parameter setting is good or not and terminate the trial early if possible.
To measure risk, we use the \vocab{Conditional Value at Risk (CVaR)}, also known as ``expected shortfall". CVaR measures the expected loss in the worst-case scenarios, defined by some quantile $\alpha$. It is computed as the expected value in the left-most tail of a distribution \citep{acerbi_expected_2002}. We use the following definition for the CVaR of a random variable $X$ for a given quantile $\alpha$: 

    \begin{equation}
    \label{cvar_equation}
        \text{CVaR}_\alpha(X) = \mathbb{E}\left[X | X \leq VaR_\alpha(X) \right]
    \end{equation}

where $\alpha \in (0, 1)$ and the $VaR_\alpha$ (Value at Risk) is just the $\alpha$-quantile of the distribution of $X$. Originally developed in finance, CVaR has also seen recent adoption in Safe RL as an additional component of the objective function by applying it to the cumulative returns within an episode, e.g. \citet{bauerle_markov_2011,chow_algorithms_2014,tamar_optimizing_2015}. In this work, we apply CVaR to the dimensions of reliability described in Section \ref{Axes of variability}.

% From the perspective of reliability, the losses will be more salient than the gains in many cases. However, there may be cases where extreme gains in performance need to be monitored. Therefore, we include measurements of the upper tails as well as the lower tails, for completeness.

\subsection{Desiderata}

In designing our metrics and statistical tests, we required that they fulfill the following criteria:
\begin{itemize}
    
    \item A minimal number of configuration parameters -- to facilitate standardization as well as to minimize ``researcher degrees of freedom" (where flexibility may allow users to tune settings to produce more favorable results, leading to an inflated rate of false positives) \citep{simmons_false-positive_2011}.
    
    \item Robust statistics, when possible. Robust statistics are less sensitive to outliers and have more reliable performance for a wider range of distributions. Robust statistics are especially important when applied to training performance, which tends to be highly non-Gaussian, making metrics such as variance and standard deviation inappropriate. For example, training performance is often bi-modal, with a concentration of points near the starting level and another concentration at the level of asymptotic performance.
    
    \item Invariance to sampling frequency -- results should not be biased by the frequency at which an algorithm was evaluated during training. See Section \ref{sampling_frequency} for further discussion.
    
    \item Enable meaningful statistical comparisons on the metrics, while making minimal assumptions about the distribution of the results. We thus designed statistical procedures that are non-parametric (Section \ref{statistics}).

\end{itemize}

\begin{figure}[!ht]
    \centering
    \begin{subfigure}[t]{0.84\textwidth}
        \includegraphics[width=\textwidth]{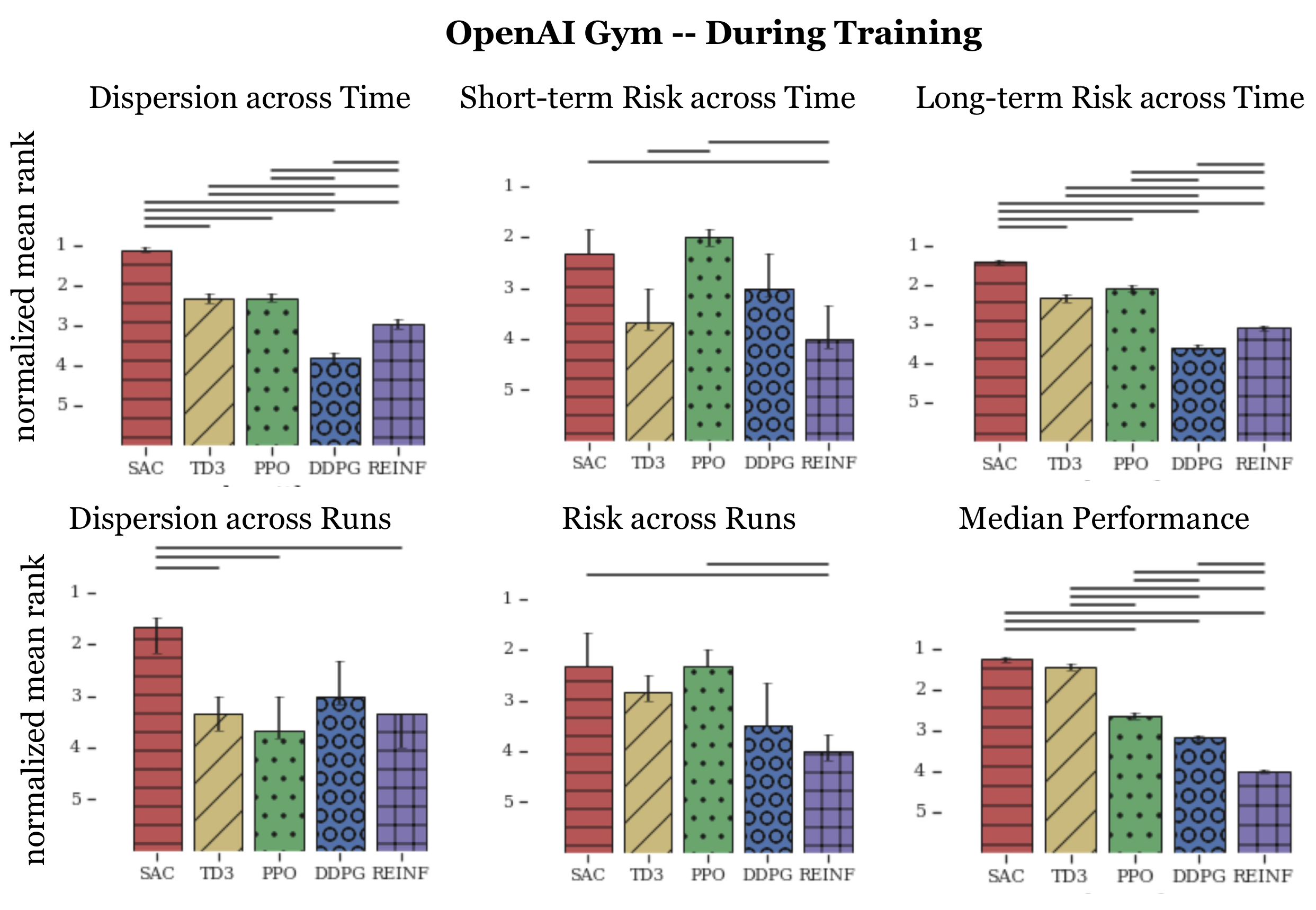}
    \end{subfigure}
    \begin{subfigure}[t]{0.84\textwidth}
        \includegraphics[width=\textwidth]{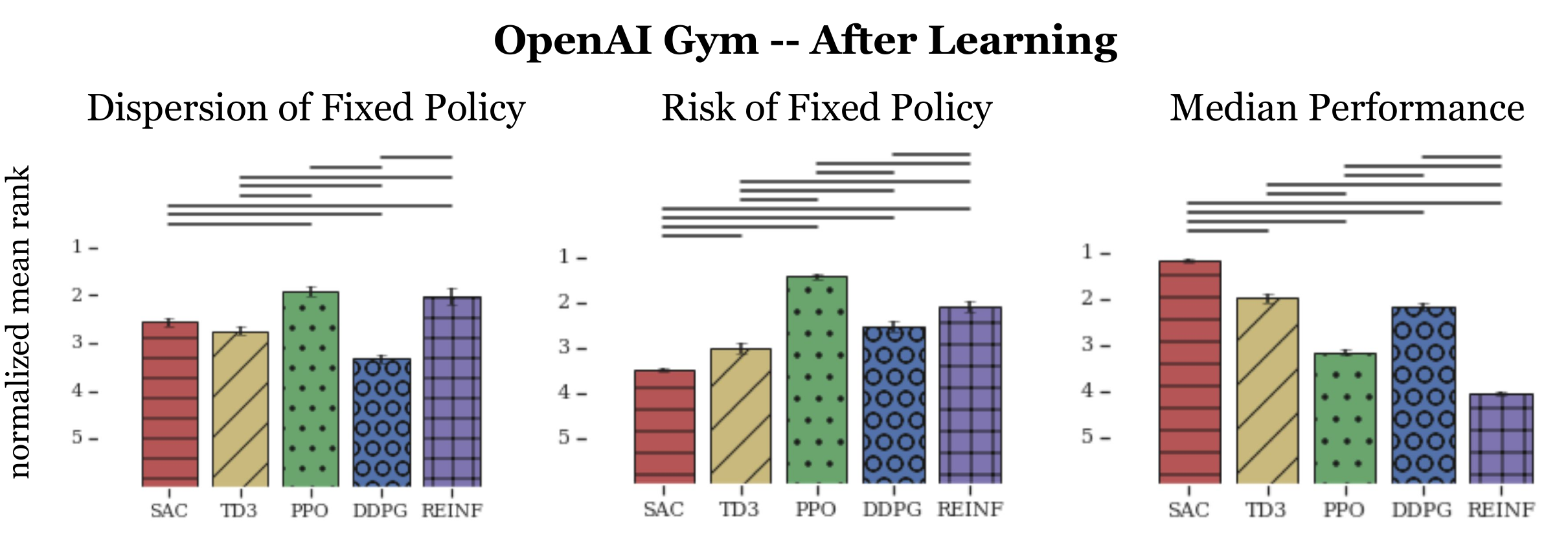}
    \end{subfigure}
    \caption{Reliability metrics and median performance for continuous control RL algorithms (DDPG, TD3, SAC, REINFORCE, and PPO) tested on OpenAI Gym environments. Rank 1 always indicates "best" reliability, e.g. lowest IQR across runs. Error bars are 95\% bootstrap confidence intervals (\# bootstraps = 1,000). Significant pairwise differences in ranking between pairs of algorithms are indicated by black horizontal lines above the colored bars. ($\alpha=0.05$ with Benjamini-Yekutieli correction, permutation test with \# permutations = 1,000). Note that the best algorithms by median performance are not always the best algorithms on reliability.}
    \label{mujoco_results_fig}
\end{figure}

\begin{figure}[!ht]
    \centering
    \begin{subfigure}[t]{0.84\textwidth}
        \includegraphics[width=\textwidth]{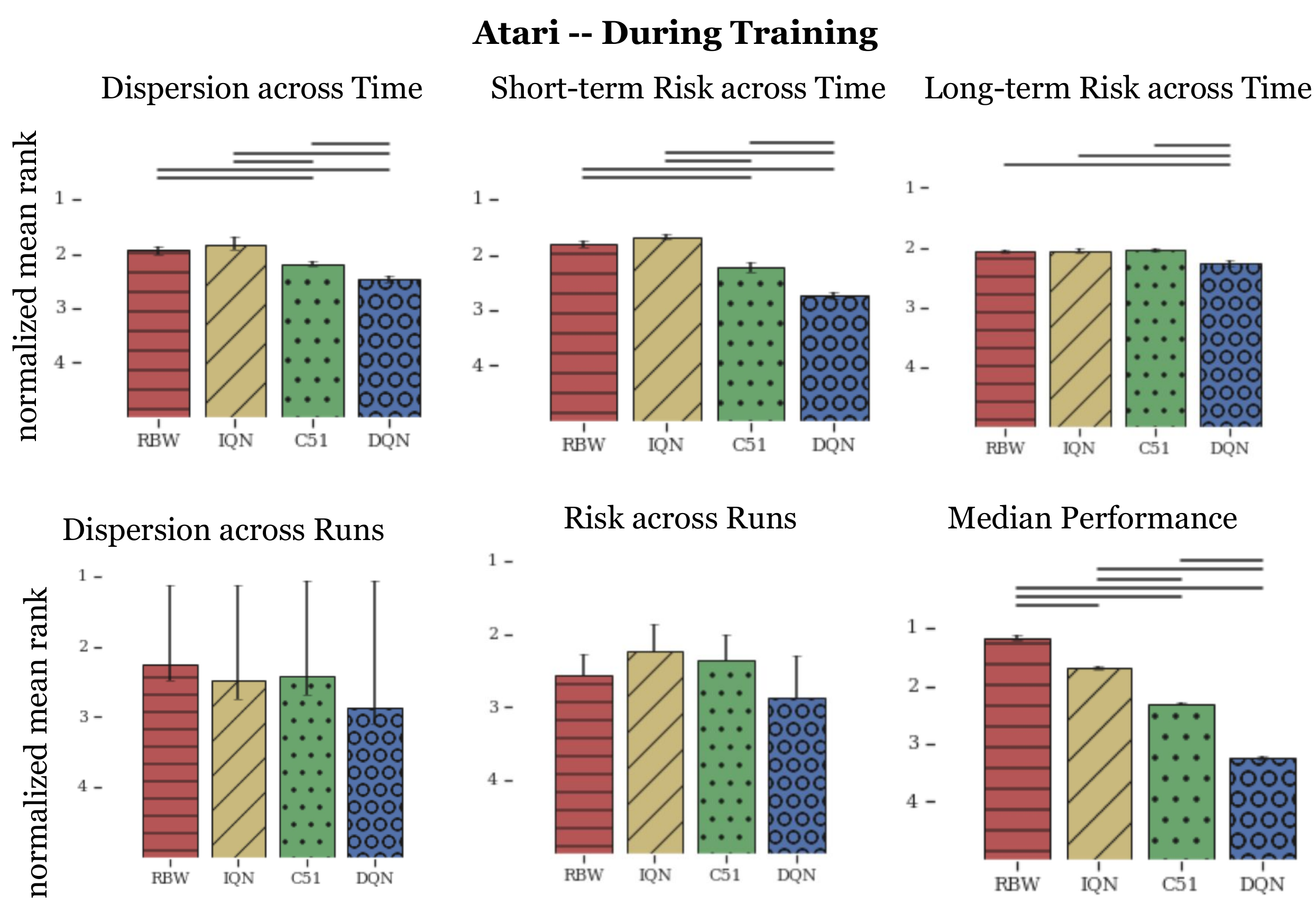}
    \end{subfigure}
    \begin{subfigure}[t]{0.84\textwidth}
        \includegraphics[width=\textwidth]{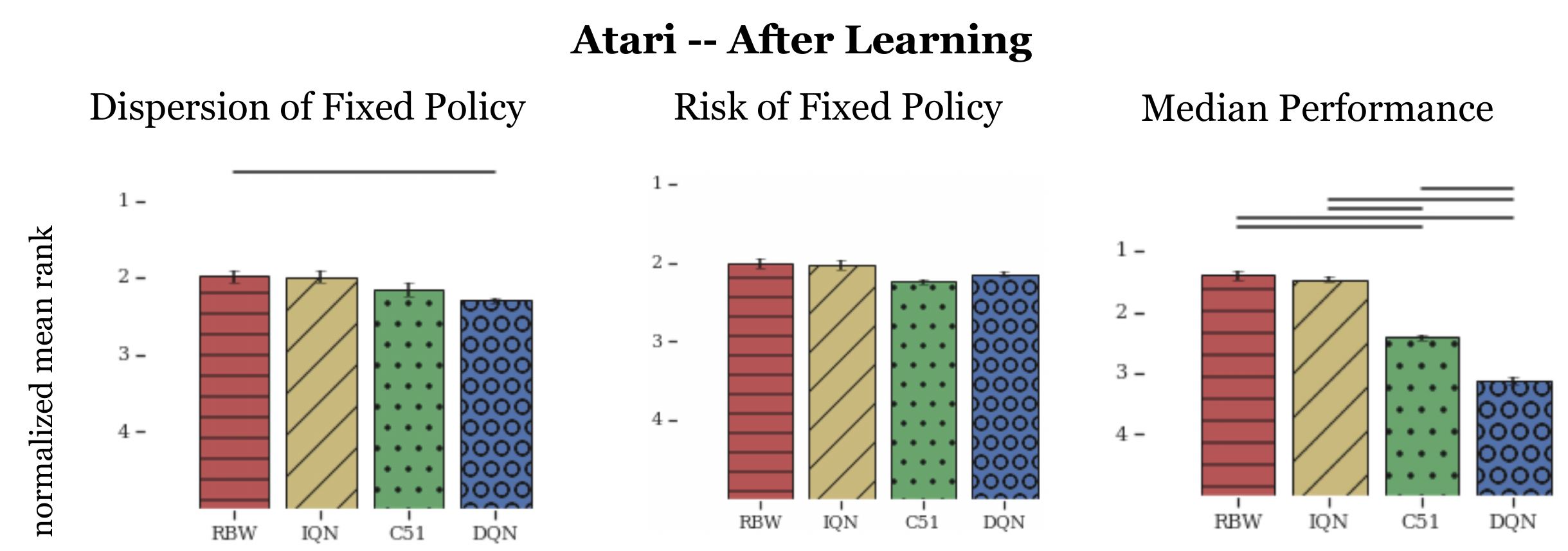}
    \end{subfigure}
    \caption{Reliability metrics and median performance for four DQN-variants (C51, DQN: Deep Q-network, IQ: Implicit Quantiles, and RBW: Rainbow) tested on 60 Atari games. Rank 1 always indicates "best" reliability, e.g. lowest IQR across runs. Significant pairwise differences in ranking between pairs of algorithms are indicated by black lines above the colored circles. ($\alpha=0.05$ with Benjamini-Yekutieli correction, permutation test with \# permutations = 1,000). Note that the best algorithms by median performance are not always the best algorithms on reliability. Error bars are 95\% bootstrap confidence intervals (\# bootstraps = 1,000).}
    \label{atari_results_fig}
\end{figure}

\subsection{Metric Definitions}

\paragraph{Dispersion across Time (DT): IQR across Time}
\label{DT}

To measure dispersion across time (DT), we wished to isolate higher-frequency variability, rather than capturing longer-term trends. We did not want our metrics to be influenced by positive trends of improvement during training, which are in fact desirable sources of variation in the training performance. Therefore, we apply detrending before computing dispersion metrics. For detrending, we used differencing (i.e. $y_t\prime = y_t - y_{t-1}$).\footnote{Please see Appendix \ref{detrending_by_differencing} for a more detailed discussion of different types of detrending, and the rationale for choosing differencing here.} The final measure consisted of inter-quartile range (IQR) within a sliding window along the detrended training curve. 

\paragraph{Short-term Risk across Time (SRT): CVaR on Differences} For this measure, we wish to measure the most extreme short-term drop over time. To do this, we apply CVaR to the changes in performance from one evaluation point to the next. I.e., in Eq. \ref{cvar_equation}, $X$ represents the differences from one evaluation time-point to the next. We first compute the time-point to time-point differences on each training run. These differences are normalized by the distance between time-points, to ensure invariance to evaluation frequency (see Section \ref{sampling_frequency}). Then, we obtain the distribution of these differences, and find the $\alpha$-quantile. Finally, we compute the expected value of the distribution below the $\alpha$-quantile. This gives us the worst-case expected drop in performance during training, from one point of evaluation to the next.

\paragraph{Long-term Risk across Time (LRT): CVaR on Drawdown} For this measure, we would also like to be able to capture whether an algorithm has the potential to lose a lot of performance relative to its peak, even if on a longer timescale, e.g. over an accumulation of small drops. For this measure, we apply CVaR to the \vocab{Drawdown}. The Drawdown at time $T$ is the drop in performance relative to the highest peak so far, and is another measure borrowed from economics \citep{chekhlov_drawdown_2005}. I.e. $\text{Drawdown}_T = R_T - \max_{t <= T}R_t $. Like the SRT metric, the LRT can capture unusually large short-term drops in performance, but can also capture unusually large drops that occur over longer timescales.

\paragraph{Dispersion across Runs (DR): IQR across Runs}
\label{DR}

Unlike the rest of the metrics described here, the dispersion across training runs has previously been used to characterize performance (e.g. \citet{duan_benchmarking_2016,islam_reproducibility_2017,bellemare_distributional_2017,fortunato_noisy_2017,nagarajan_deterministic_2018}). This is usually measured by taking the variance or standard deviation across training runs at a set of evaluation points. 
%As the RL community greatly benefits from this kind of information, we advocate reporting it more consistently. 
We build on the existing practice by recommending first performing low-pass filtering of the training data, to filter out high-frequency variability within runs (this is instead measured using Dispersion across Time, DT). We also replace variance or standard deviation with robust statistics like IQR.

\paragraph{Risk across Runs (RR): CVaR across Runs}

In order to measure Risk across Runs (RR), we apply CVaR to the final performance of all the training runs. This gives a measure of the expected performance of the worst runs.
% For example, on the well-known Hopper benchmark task, this measure might capture the average performance runs that perform poorly at discovering the proper gait.

\paragraph{Dispersion across Fixed-Policy Rollouts (DF): IQR across Rollouts}

When evaluating a fixed policy, we are interested in variability in performance when the same policy is rolled out multiple times. To compute this metric, we simply compute the IQR on the performance of the rollouts.

\paragraph{Risk across Fixed-Policy Rollouts (RF): CVaR across Rollouts}

This metric is similar to DF, except that we apply CVaR on the rollout performances.

\subsection{Invariance to frequency of evaluation}
\label{sampling_frequency}

Different experiments and different tasks may produce evaluations at different frequencies during training. Therefore, the reliability metrics should be unbiased by the choice of evaluation frequency. As long as there are no cyclical patterns in performance, the frequency of evaluation will not bias any of the metrics except Long-Term Risk across Time (LRT). For all other metrics, changes in the frequency of evaluation will simply lead to more or less noisy estimates of these metrics. For LRT, comparisons should only be made if the frequency of evaluation is held constant across experiments.

% Of all the metrics defined here, only Long-Term Risk Across Time (LRT) is affected by the frequency of evaluation. With a lower frequency, the drawdown is smaller on average, because there is a chance that the peaks/troughs are removed. This should be taken into account when making comparisons, i.e. comparisons of LRT should only be done if the frequency of evaluation is held constant across experiments. For more details, see Section \ref{reporting}.

\section{Recommendations for reporting metrics and parameters}
\label{reporting}

Whether evaluating an algorithm for practical use or for research, we recommend evaluating all of the reliability metrics described above. Each metric measures a different aspect of reliability, and can help pinpoint specific strengths and weaknesses of the algorithm. Evaluating the metrics is easy with the open-source Python package that we have released.

\paragraph{Reporting parameters.} Even given our purposeful efforts to minimize the number of parameters in the reliability metrics, a few remain to be specified by the user that can affect the results, namely: window size (for Dispersion across Time), frequency threshold for low-pass and high-pass filtering (Dispersion across Time, Dispersion across Runs), evaluation frequency (only for Long-term Risk across Time), and length of training runs. Therefore, when reporting these metrics, these parameters need to be clearly specified, and must also be held constant across experiments for meaningful comparisons. The same is true for any other parameters that affect evaluation, e.g., the number of roll-outs per evaluation, the parameters of the environment, whether on-line or off-line evaluation is used, and the random seeds chosen.

\paragraph{Collapsing across evaluation points.} Some of the in-training reliability metrics (Dispersion across Runs, Risk across Runs, and Dispersion across Time) need to be evaluated at multiple evaluation points along the training runs. If it is useful to obtain a small number of values to summarize each metric, we recommend dividing the training run into "time frames" (e.g. beginning, middle, and end), and collapsing across all evaluation points within each time frame.

\paragraph{Normalization by performance.} Different algorithms can have vastly different ranges of performance even on the same task, and variability in performance tends to scale with actual performance. Thus, we normalize our metrics in post-processing by a measure of the range of performance for each algorithm. For "during training" reliability, we recommend normalizing by the median range of performance, which we define as the $p_{P_95} - p_{t=0}$, where $p_{P_95}$ is the 95th percentile and $p_{t=0}$ is the starting performance. For "after learning" reliability, the range of performance may not be available, in which case we use the median performance directly.

\paragraph{Ranking the algorithms.} Because different environments have different ranges and distributions of reward, we must be careful when aggregating across environments or comparing between environments. Thus, if the analysis involves more than one environment, the per-environment median results for the algorithms are first converted to rankings, by ranking all algorithms within each task. To summarize the performance of a single algorithm across multiple tasks, we compute the mean ranking across tasks.

\paragraph{Per-environment analysis.} The same algorithm can have different patterns of reliability for different environments. Therefore, we recommend inspecting reliability metrics on a per-environment basis, as well as aggregating across environments as described above.

\section{Confidence Intervals and Statistical Significance Tests for Comparison}
\label{statistics}

\subsection{Confidence intervals}

We assume that the metric values have been converted to mean rankings, as explained in Section \ref{reporting}. To obtain confidence intervals on the mean rankings for each algorithm, we apply bootstrap sampling on the runs, by resampling runs with replacement \citep{efron_bootstrap_1986}.

For metrics that are evaluated per-run (e.g. Dispersion across Time), we can resample the metric values directly, and then recompute the mean rankings on each resampling to obtain a distribution over the rankings; this allow us to compute confidence intervals. For metrics that are evaluated across-runs, we need to resample the runs themselves, then evaluate the metrics on each resampling, before recomputing the mean rankings to obtain a distribution on the mean rankings.

\subsection{Significance Tests for Comparing algorithms}

Commonly, we would like to compare algorithms evaluated on a fixed set of environments. To determine whether any two algorithms have statistically significant differences in their metric rankings, we perform an exact permutation test on each pair of algorithms. Such tests allow us to compute a p-value for the null hypothesis (probability that the methods are in fact indistinguishable on the reliability metric).

We designed our permutation tests based on the null hypothesis that runs are exchangeable across the two algorithms being compared. In brief, let $A$ and $B$ be sets of performance measurements for algorithms $a$ and $b$. Let $Metric(X)$ be a reliability metric, e.g. the inter-quartile range across runs, computed on a set of measurements $X$. $MetricRanking(X)$ is the mean ranking across tasks on $X$, compared to the other algorithms being considered. We compute test statistic $$s_{MetricRanking}(A,B)=MetricRanking(A)-MetricRanking(B).$$ Next we compute the distribution for $s_{MetricRanking}$ under the null hypothesis that the methods are equivalent, i.e. that performance measurements should have the same distribution for $a$ and $b$. We do this by computing random partitions $A',B'$ of $\{A \cup B\}$, and computing the test statistic $s_{MetricRanking}(A',B')$ on each partition. This yields a distribution for $s_{MetricRanking}$ (for sufficiently many samples), and the p-value can be computed from the percentile value of $s_{MetricRanking}(A,B)$ in this distribution. As with the confidence intervals, a different procedure is required for per-run vs across-run metrics. Please see Appendix \ref{permutation_test_illustrations} for diagrams illustrating the permutation test procedures.

When performing pairwise comparisons between algorithms, it is critical to include corrections for multiple comparisons. This is because the probability of incorrect inferences increases with a greater number of simultaneous comparisons. We recommend using the Benjamini-Yekutieli method, which controls the false discovery rate (FDR), i.e., the proportion of rejected null hypotheses that are false.\footnote{For situations in which a user wishes instead to control the family-wise error rate (FWER; the probability of incorrectly rejecting at least one true null hypothesis), we recommend using the Holm-Bonferroni method.}

% The most straightforward (but most stringent) is the Bonferroni correction, which controls the familywise error rate (FWER), i.e. the probability of incorrectly rejecting at least one true null hypothesis.\footnote{Other options include controlling the false discovery rate (FDR) instead of the FWER.} To control the FWER at $p \leq \alpha$, we adjust the significance threshold for each paired comparison to be $\alpha / m$, where $m$ is the number of comparisons. 
%For example, for an uncorrected significance threshold of 0.05 and four algorithms, we should use a corrected threshold of $0.05 / \binom{4}{2} = 0.05/6$.

\subsection{Reporting on statistical tests}

It is important to report the details of any statistical tests performed, e.g. which test was used, the significance threshold, and the type of multiple-comparisons correction used.

\section{Analysis of Reliability for Common Algorithms and Environments}
\label{analysis}

In this section, we provide examples of applying the reliability metrics to a number of RL algorithms and environments, following the recommendations described above.

\subsection{Continuous control algorithms on OpenAI Gym}
\label{MuJoCo_data}

We applied the reliability metrics to algorithms tested on seven continuous control environments from the Open-AI Gym \citep{greg_brockman_openai_2016} run on the MuJoCo physics simulator \citep{todorov_MuJoCo:_2012}. We tested REINFORCE \citep{sutton_policy_2000}, DDPG \citep{lillicrap_continuous_2015}, PPO \citep{schulman_proximal_2017}, TD3 \citep{fujimoto_addressing_2018}, and SAC \citep{haarnoja_soft_2018} on the following Gym environments: Ant-v2, HalfCheetah-v2, 
% Hopper-v2, 
Humanoid-v2, Reacher-v2, Swimmer-v2, and Walker2d-v2. We used the implementations of DDPG, TD3, and SAC from the TF-Agents library \citep{TFAgents}. Each algorithm was run on each environment for 30 independent training runs.

We used a black-box optimizer \citep{golovin_google_2017} to tune selected hyperparameters on a per-task basis, optimizing for final performance.
The remaining hyperparameters were defined as stated in the corresponding original papers.
See Appendix \ref{hyperparameters} for details of the hyperparameter search space and the final set of hyperparameters.
% \footnote{See Appendix \ref{vizier_trials} for an analysis of reliability across hyperparameters, based on the data generated during this black-box optimization.}
During training, we evaluated the policies at a frequency of 1000 training steps. Each algorithm was run for a total of two million environment steps. For the ``online'' evaluations we used the generated training curves, averaging returns over recent training episodes collected using the exploration policy as it evolves. The raw training curves are shown in Appendix \ref{raw_mujoco_curves}. For evaluations after learning on a fixed policy, we took the last checkpoint from each training run as the fixed policy for evaluation. Each of these policies was then evaluated for 30 roll-outs, where each roll-out was defined as 1000 environment steps.

\subsection{Discrete control: DQN variants on Atari}
\label{atari_data}

We also applied the reliability metrics to the RL algorithms and training data released as part of the Dopamine package \citep{dopamine}. The data comprise the training runs of four RL algorithms, each applied to 60 Atari games. The RL algorithms are: DQN \citep{mnih_human-level_2015}, Implicit Quantile (IQN) \citep{dabney_implicit_2018}, C51 \citep{bellemare_distributional_2017}, and a variant of Rainbow implementing the three most important components \citep{hessel_rainbow:_2018}. 
%Implicit Quantile, C51, and Rainbow are all extensions of DQN that were designed to improve its performance.
The algorithms were trained on each game for 5 training runs. Hyper-parameters follow the original papers, but were modified as necessary to follow Rainbow \citep{hessel_rainbow:_2018}, to ensure apples-to-apples comparison. See Appendix \ref{hyperparameters} for the hyperparameters.

During training, the algorithms were evaluated in an ``online'' fashion every 1 million frames, averaging across the training episodes as recommended for evaluations on the ALE \citep{machado_revisiting_2018}. Each training run consisted of approximately 200 million Atari frames (rounding to the nearest episode boundary every 1 million frames).\footnote{The raw training curves can be viewed at https://google.github.io/dopamine/baselines/plots.html} For evaluations after learning on a fixed policy (``after learning''), we took the last checkpoint from each training run as the fixed policies for evaluation. We then evaluated each of these policies for 125,000 environment steps. 
%One episode / rollout was defined following the standard definition for each game.

% vizier is based off the evals, but I'm reporting intermediate results as well (evaluation runs on the policy as it trains). only the last one should actually be used by vizier though. the train metrics are computed over the experience generated during the collect phase of training, so they're using the exploration policy. the eval runs always use the eval policy (greedy)

\subsection{Parameters for reliability metrics, confidence intervals, and statistical tests}

For the MuJoCo environments, we applied a sliding window of 100000 training steps for Dispersion across Time. For the Atari experiments, we used a sliding window size of 25 on top of the evaluations for the Dispersion across Time. For metrics with multiple evaluation points, we divided each training run into 3 time frames and averaged the metric rankings within each time frame. Because the results were extremely similar for all three time frames, we here report just for the final time frames.

Statistical tests for comparing algorithms were performed according to the recommendations in Section \ref{statistics}. We used pairwise permutation tests using 10,000 permutations per test, with a significance threshold of 0.05 and Benjamini-Yekutieli multiple-comparisons correction.

\subsection{Median performance}

The median performance of an algorithm is not a reliability metric, but it is interesting to see side-by-side with the reliability metrics. For analyzing median performance for the DQN variants, we used the normalization scheme of \citep{mnih_human-level_2015}, where an algorithm's performance is normalized against a lower baseline (e.g. the performance of a random policy) and an upper baseline (e.g. the performance of a human): $P_\text{normalized} = \frac{P - B_\text{lower}}{B_\text{upper} - B_\text{lower}}$. Median performance was not normalized for the continuous control algorithms. 

\subsection{Results}

The reliability metric rankings are shown in Fig. \ref{mujoco_results_fig} for the MuJoCo results. We see that, according to Median Performance during training, SAC and TD3 have the best performance and perform similarly well, while REINFORCE performs the worst. However, SAC outperforms TD3 on all reliability metrics during training. Furthermore, both SAC and TD3 perform relatively poorly on all reliability metrics after learning, despite performing best on median performance.

The reliability metric rankings are shown in Fig. \ref{atari_results_fig} for the Atari results. Here we see a similar result that, even though Rainbow performs significantly better than IQN in Median Performance, IQN performs numerically or significantly better than Rainbow on many of the reliability metrics.

The differing patterns in these metrics demonstrates that reliability is a separate dimension that needs to be inspected separately from mean or median performance -- two algorithms may have similar median performance but may nonetheless significantly differ in reliability, as with SAC and TD3 above. Additionally, these results demonstrate that reliability along one axis does not necessarily correlate with reliability on other axes, demonstrating the value of evaluating these different dimensions so that algorithms can be compared and selected based on the requirements of the problem at hand.

To see metric results evaluated on a per-environment basis, please refer to Appendix \ref{per_task_results}. Rank order of algorithms was often relatively consistent across the different environments evaluated. However, different environments did display different patterns across algorithms. For example, even though SAC showed the same or better Dispersion across Runs for most of the MuJoCo environments evaluated, it did show slightly worse Dispersion across Runs for the HalfCheetah environment (Fig  \ref{mujoco_per_task_DR}). This kind of result emphasizes the importance of inspecting reliability (and other performance metrics) on a per-environment basis, and also of evaluating reliability and performance on the environment of interest, if possible.

\section{Conclusion}

We have presented a number of metrics, designed to measure different aspects of reliability of RL algorithms. We motivated the design goals and choices made in constructing these metrics, and also presented practical recommendations for the measurement of reliability for RL. Additionally, we presented examples of applying these metrics to common RL algorithms and environments, and showed that these metrics can reveal strengths and weaknesses of an algorithm that are obscured when we only inspect mean or median performance.
% We recommend evaluating RL algorithms on such metrics, whether for research or production, and provide an open-source package to facilitate it.

% \subsubsection*{Author Contributions}
% If you'd like to, you may include  a section for author contributions as is done
% in many journals. This is optional and at the discretion of the authors.

\subsubsection*{Acknowledgments}

Many thanks to the following people for helpful discussions during the formulation of these metrics and the writing of the paper: Mohammad Ghavamzadeh, Yinlam Chow, Danijar Hafner, Rohan Anil, Archit Sharma, Vikas Sindhwani, Krzysztof Choromanski, Joelle Pineau, Hal Varian, Shyue-Ming Loh, and Tim Hesterberg. Thanks also to Toby Boyd for his assistance in the open-sourcing process, Oscar Ramirez for code reviews, and Pablo Castro for his help with running experiments using the Dopamine baselines data.

\bibliography{refs_zotero,refs_manual}

\begin{thebibliography}{33}
\providecommand{\natexlab}[1]{#1}
\providecommand{\url}[1]{\texttt{#1}}
\expandafter\ifx\csname urlstyle\endcsname\relax
  \providecommand{\doi}[1]{doi: #1}\else
  \providecommand{\doi}{doi: \begingroup \urlstyle{rm}\Url}\fi

\bibitem[Acerbi \& Tasche(2002)Acerbi and Tasche]{acerbi_expected_2002}
Carlo Acerbi and Dirk Tasche.
\newblock Expected {Shortfall}: {A} {Natural} {Coherent} {Alternative} to
  {Value} at {Risk}.
\newblock \emph{Economic Notes}, 31\penalty0 (2):\penalty0 379--388, July 2002.
\newblock ISSN 0391-5026, 1468-0300.
\newblock \doi{10.1111/1468-0300.00091}.
\newblock URL \url{http://doi.wiley.com/10.1111/1468-0300.00091}.

\bibitem[Bellemare et~al.(2017)Bellemare, Dabney, and
  Munos]{bellemare_distributional_2017}
Marc~G. Bellemare, Will Dabney, and Rémi Munos.
\newblock A {Distributional} {Perspective} on {Reinforcement} {Learning}.
\newblock \emph{arXiv:1707.06887 [cs, stat]}, July 2017.
\newblock URL \url{http://arxiv.org/abs/1707.06887}.
\newblock arXiv: 1707.06887.

\bibitem[Bäuerle \& Ott(2011)Bäuerle and Ott]{bauerle_markov_2011}
Nicole Bäuerle and Jonathan Ott.
\newblock Markov {Decision} {Processes} with {Average}-{Value}-at-{Risk}
  criteria.
\newblock \emph{Mathematical Methods of Operations Research}, 74\penalty0
  (3):\penalty0 361--379, December 2011.
\newblock ISSN 1432-2994, 1432-5217.
\newblock \doi{10.1007/s00186-011-0367-0}.
\newblock URL \url{http://link.springer.com/10.1007/s00186-011-0367-0}.

\bibitem[Castro et~al.(2018)Castro, Moitra, Gelada, Kumar, and
  Bellemare]{dopamine}
Pablo~Samuel Castro, Subhodeep Moitra, Carles Gelada, Saurabh Kumar, and
  Marc~G. Bellemare.
\newblock Dopamine: {A} research framework for deep reinforcement learning.
\newblock \emph{CoRR}, abs/1812.06110, 2018.
\newblock URL \url{http://arxiv.org/abs/1812.06110}.

\bibitem[Chekhlov et~al.(2005)Chekhlov, Uryasev, and
  Zabarankin]{chekhlov_drawdown_2005}
Alexei Chekhlov, Stanislav Uryasev, and Michael Zabarankin.
\newblock Drawdown measure in portfolio optimization.
\newblock \emph{International Journal of Theoretical and Applied Finance},
  8\penalty0 (1):\penalty0 46, 2005.

\bibitem[Chow \& Ghavamzadeh(2014)Chow and Ghavamzadeh]{chow_algorithms_2014}
Yinlam Chow and Mohammad Ghavamzadeh.
\newblock Algorithms for {CVaR} {Optimization} in {MDPs}.
\newblock \emph{Advances in Neural Information Processing Systems}, pp.\ ~9,
  2014.

\bibitem[Colas et~al.(2018)Colas, Sigaud, and Oudeyer]{colas_how_2018}
Cédric Colas, Olivier Sigaud, and Pierre-Yves Oudeyer.
\newblock How {Many} {Random} {Seeds}? {Statistical} {Power} {Analysis} in
  {Deep} {Reinforcement} {Learning} {Experiments}.
\newblock \emph{arXiv:1806.08295 [cs, stat]}, June 2018.
\newblock URL \url{http://arxiv.org/abs/1806.08295}.
\newblock arXiv: 1806.08295.

\bibitem[Colas et~al.(2019)Colas, Sigaud, and Oudeyer]{colas_hitchhikers_2019}
Cédric Colas, Olivier Sigaud, and Pierre-Yves Oudeyer.
\newblock A {Hitchhiker}'s {Guide} to {Statistical} {Comparisons} of
  {Reinforcement} {Learning} {Algorithms}.
\newblock \emph{arXiv:1904.06979 [cs, stat]}, April 2019.
\newblock URL \url{http://arxiv.org/abs/1904.06979}.
\newblock arXiv: 1904.06979.

\bibitem[Dabney et~al.(2018)Dabney, Ostrovski, Silver, and
  Munos]{dabney_implicit_2018}
Will Dabney, Georg Ostrovski, David Silver, and Rémi Munos.
\newblock Implicit {Quantile} {Networks} for {Distributional} {Reinforcement}
  {Learning}.
\newblock \emph{Thirty-fith International Conference on Machine Learning}, pp.\
  ~10, 2018.

\bibitem[Duan et~al.(2016)Duan, Chen, Houthooft, Schulman, and
  Abbeel]{duan_benchmarking_2016}
Yan Duan, Xi~Chen, Rein Houthooft, John Schulman, and Pieter Abbeel.
\newblock Benchmarking {Deep} {Reinforcement} {Learning} for {Continuous}
  {Control}.
\newblock In \emph{International {Conference} on {Machine} {Learning}}, pp.\
  1329--1338, June 2016.
\newblock URL \url{http://proceedings.mlr.press/v48/duan16.html}.

\bibitem[Efron \& Tibshirani(1986)Efron and Tibshirani]{efron_bootstrap_1986}
B.~Efron and R.~Tibshirani.
\newblock Bootstrap {Methods} for {Standard} {Errors}, {Confidence}
  {Intervals}, and {Other} {Measures} of {Statistical} {Accuracy}.
\newblock \emph{Statistical Science}, 1\penalty0 (1):\penalty0 54--75, February
  1986.
\newblock ISSN 0883-4237, 2168-8745.
\newblock \doi{10.1214/ss/1177013815}.
\newblock URL \url{http://projecteuclid.org/euclid.ss/1177013815}.

\bibitem[Fortunato et~al.(2017)Fortunato, Azar, Piot, Menick, Osband, Graves,
  Mnih, Munos, Hassabis, Pietquin, Blundell, and Legg]{fortunato_noisy_2017}
Meire Fortunato, Mohammad~Gheshlaghi Azar, Bilal Piot, Jacob Menick, Ian
  Osband, Alex Graves, Vlad Mnih, Remi Munos, Demis Hassabis, Olivier Pietquin,
  Charles Blundell, and Shane Legg.
\newblock Noisy {Networks} for {Exploration}.
\newblock \emph{arXiv:1706.10295 [cs, stat]}, June 2017.
\newblock URL \url{http://arxiv.org/abs/1706.10295}.
\newblock arXiv: 1706.10295.

\bibitem[Fujimoto et~al.(2018)Fujimoto, van Hoof, and
  Meger]{fujimoto_addressing_2018}
Scott Fujimoto, Herke van Hoof, and David Meger.
\newblock Addressing {Function} {Approximation} {Error} in {Actor}-{Critic}
  {Methods}.
\newblock \emph{arXiv:1802.09477 [cs, stat]}, February 2018.
\newblock URL \url{http://arxiv.org/abs/1802.09477}.
\newblock arXiv: 1802.09477.

\bibitem[Golovin et~al.(2017)Golovin, Solnik, Moitra, Kochanski, Karro, and
  Sculley]{golovin_google_2017}
Daniel Golovin, Benjamin Solnik, Subhodeep Moitra, Greg Kochanski, John Karro,
  and D.~Sculley.
\newblock Google {Vizier}: {A} {Service} for {Black}-{Box} {Optimization}.
\newblock In \emph{Proceedings of the 23rd {ACM} {SIGKDD} {International}
  {Conference} on {Knowledge} {Discovery} and {Data} {Mining} - {KDD} '17},
  pp.\  1487--1495, Halifax, NS, Canada, 2017. ACM Press.
\newblock ISBN 978-1-4503-4887-4.
\newblock \doi{10.1145/3097983.3098043}.
\newblock URL \url{http://dl.acm.org/citation.cfm?doid=3097983.3098043}.

\bibitem[{Greg Brockman} et~al.(2016){Greg Brockman}, {Vicki Cheung}, {Ludwig
  Pettersson}, {Jonas Schneider}, {John Schulman}, {Jie Tang}, and {Wojciech
  Zaremba}]{greg_brockman_openai_2016}
{Greg Brockman}, {Vicki Cheung}, {Ludwig Pettersson}, {Jonas Schneider}, {John
  Schulman}, {Jie Tang}, and {Wojciech Zaremba}.
\newblock {OpenAI} {Gym}, 2016.

\bibitem[Guadarrama et~al.(2018)Guadarrama, Korattikara, Oscar~Ramirez, Holly,
  Fishman, Wang, Ekaterina~Gonina, Vanhoucke, and Brevdo]{TFAgents}
Sergio Guadarrama, Anoop Korattikara, Pablo~Castro Oscar~Ramirez, Ethan Holly,
  Sam Fishman, Ke~Wang, Chris~Harris Ekaterina~Gonina, Vincent Vanhoucke, and
  Eugene Brevdo.
\newblock {TF-Agents}: A library for reinforcement learning in tensorflow.
\newblock \url{https://github.com/tensorflow/agents}, 2018.
\newblock URL \url{https://github.com/tensorflow/agents}.
\newblock [Online; accessed 30-November-2018].

\bibitem[Haarnoja et~al.(2018)Haarnoja, Zhou, Abbeel, and
  Levine]{haarnoja_soft_2018}
Tuomas Haarnoja, Aurick Zhou, Pieter Abbeel, and Sergey Levine.
\newblock Soft {Actor}-{Critic}: {Off}-{Policy} {Maximum} {Entropy} {Deep}
  {Reinforcement} {Learning} with a {Stochastic} {Actor}.
\newblock \emph{arXiv:1801.01290 [cs, stat]}, January 2018.
\newblock URL \url{http://arxiv.org/abs/1801.01290}.
\newblock arXiv: 1801.01290.

\bibitem[Hamilton(1994)]{hamilton_time_1994}
James~D. Hamilton.
\newblock \emph{Time {Series} {Analysis}}.
\newblock Princeton University Press, 1994.

\bibitem[Henderson et~al.(2017)Henderson, Islam, Bachman, Pineau, Precup, and
  Meger]{henderson_deep_2017}
Peter Henderson, Riashat Islam, Philip Bachman, Joelle Pineau, Doina Precup,
  and David Meger.
\newblock Deep {Reinforcement} {Learning} that {Matters}.
\newblock \emph{arXiv:1709.06560 [cs, stat]}, September 2017.
\newblock URL \url{http://arxiv.org/abs/1709.06560}.
\newblock arXiv: 1709.06560.

\bibitem[Hessel \& Modayil(2018)Hessel and Modayil]{hessel_rainbow:_2018}
Matteo Hessel and Joseph Modayil.
\newblock Rainbow: {Combining} {Improvements} in {Deep} {Reinforcement}
  {Learning}.
\newblock \emph{AAAI}, pp.\ ~8, 2018.

\bibitem[Islam et~al.(2017)Islam, Henderson, Gomrokchi, and
  Precup]{islam_reproducibility_2017}
Riashat Islam, Peter Henderson, Maziar Gomrokchi, and Doina Precup.
\newblock Reproducibility of {Benchmarked} {Deep} {Reinforcement} {Learning}
  {Tasks} for {Continuous} {Control}.
\newblock \emph{arXiv:1708.04133 [cs]}, August 2017.
\newblock URL \url{http://arxiv.org/abs/1708.04133}.
\newblock arXiv: 1708.04133.

\bibitem[Lillicrap et~al.(2015)Lillicrap, Hunt, Pritzel, Heess, Erez, Tassa,
  Silver, and Wierstra]{lillicrap_continuous_2015}
Timothy~P. Lillicrap, Jonathan~J. Hunt, Alexander Pritzel, Nicolas Heess, Tom
  Erez, Yuval Tassa, David Silver, and Daan Wierstra.
\newblock Continuous control with deep reinforcement learning.
\newblock \emph{arXiv:1509.02971 [cs, stat]}, September 2015.
\newblock URL \url{http://arxiv.org/abs/1509.02971}.
\newblock arXiv: 1509.02971.

\bibitem[Machado et~al.(2018)Machado, Bellemare, Talvitie, Veness, Hausknecht,
  and Bowling]{machado_revisiting_2018}
Marlos~C. Machado, Marc~G. Bellemare, Erik Talvitie, Joel Veness, Matthew
  Hausknecht, and Michael Bowling.
\newblock Revisiting the {Arcade} {Learning} {Environment}: {Evaluation}
  {Protocols} and {Open} {Problems} for {General} {Agents}.
\newblock \emph{Journal of Artificial Intelligence Research}, 61:\penalty0
  523--562, March 2018.
\newblock ISSN 1076-9757.
\newblock \doi{10.1613/jair.5699}.
\newblock URL \url{https://www.jair.org/index.php/jair/article/view/11182}.

\bibitem[Mnih et~al.(2015)Mnih, Kavukcuoglu, Silver, Rusu, Veness, Bellemare,
  Graves, Riedmiller, Fidjeland, Ostrovski, Petersen, Beattie, Sadik,
  Antonoglou, King, Kumaran, Wierstra, Legg, and
  Hassabis]{mnih_human-level_2015}
Volodymyr Mnih, Koray Kavukcuoglu, David Silver, Andrei~A. Rusu, Joel Veness,
  Marc~G. Bellemare, Alex Graves, Martin Riedmiller, Andreas~K. Fidjeland,
  Georg Ostrovski, Stig Petersen, Charles Beattie, Amir Sadik, Ioannis
  Antonoglou, Helen King, Dharshan Kumaran, Daan Wierstra, Shane Legg, and
  Demis Hassabis.
\newblock Human-level control through deep reinforcement learning.
\newblock \emph{Nature}, 518\penalty0 (7540):\penalty0 529--533, February 2015.
\newblock ISSN 1476-4687.
\newblock \doi{10.1038/nature14236}.
\newblock URL \url{https://www.nature.com/articles/nature14236/}.

\bibitem[Nagarajan et~al.(2018)Nagarajan, Warnell, and
  Stone]{nagarajan_deterministic_2018}
Prabhat Nagarajan, Garrett Warnell, and Peter Stone.
\newblock Deterministic {Implementations} for {Reproducibility} in {Deep}
  {Reinforcement} {Learning}.
\newblock \emph{arXiv:1809.05676 [cs]}, September 2018.
\newblock URL \url{http://arxiv.org/abs/1809.05676}.
\newblock arXiv: 1809.05676.

\bibitem[Nelson \& Plosser(1982)Nelson and Plosser]{nelson_trends_1982}
Charles~R Nelson and Charles~I Plosser.
\newblock Trends and random walks in macroeconomic time series.
\newblock \emph{Journal of Monetary Economics}, 10:\penalty0 139--162, 1982.

\bibitem[Rousseeuw \& Croux(1993)Rousseeuw and
  Croux]{peter_j._rousseeuw_alternatives_1993}
Peter~J. Rousseeuw and Christophe Croux.
\newblock Alternatives to the {MedianAbsolute} {Deviation}.
\newblock \emph{Journal of the American Statistical Association}, 1993.

\bibitem[{Said E. Said} \& {David A. Dickey}(1984){Said E. Said} and {David A.
  Dickey}]{said_e._said_testing_1984}
{Said E. Said} and {David A. Dickey}.
\newblock Testing for unit roots in autoregressive-moving average models of
  unknown order.
\newblock \emph{Biometrika}, 71\penalty0 (3):\penalty0 599--607, 1984.

\bibitem[Schulman et~al.(2017)Schulman, Wolski, Dhariwal, Radford, and
  Klimov]{schulman_proximal_2017}
John Schulman, Filip Wolski, Prafulla Dhariwal, Alec Radford, and Oleg Klimov.
\newblock Proximal {Policy} {Optimization} {Algorithms}.
\newblock \emph{arXiv:1707.06347 [cs]}, July 2017.
\newblock URL \url{http://arxiv.org/abs/1707.06347}.
\newblock arXiv: 1707.06347.

\bibitem[Simmons et~al.(2011)Simmons, Nelson, and
  Simonsohn]{simmons_false-positive_2011}
Joseph~P. Simmons, Leif~D. Nelson, and Uri Simonsohn.
\newblock False-{Positive} {Psychology}: {Undisclosed} {Flexibility} in {Data}
  {Collection} and {Analysis} {Allows} {Presenting} {Anything} as
  {Significant}.
\newblock \emph{Psychological Science}, 22\penalty0 (11):\penalty0 1359--1366,
  November 2011.
\newblock ISSN 0956-7976, 1467-9280.
\newblock \doi{10.1177/0956797611417632}.
\newblock URL \url{http://journals.sagepub.com/doi/10.1177/0956797611417632}.

\bibitem[Sutton et~al.(2000)Sutton, McAllester, Singh, and
  Mansour]{sutton_policy_2000}
Richard~S Sutton, David~A McAllester, Satinder~P Singh, and Yishay Mansour.
\newblock Policy {Gradient} {Methods} for {Reinforcement} {Learning} with
  {Function} {Approximation}.
\newblock In \emph{{NIPS}'99 {Proceedings} of the 12th {International}
  {Conference} on {Neural} {Information} {Processing} {Systems}}, pp.\ ~7,
  2000.

\bibitem[Tamar et~al.(2015)Tamar, Glassner, and Mannor]{tamar_optimizing_2015}
Aviv Tamar, Yonatan Glassner, and Shie Mannor.
\newblock Optimizing the {CVaR} via {Sampling}.
\newblock \emph{Proceedings of the Twenty-Ninth AAAI Conference on Artificial
  Intelligence}, pp.\ ~7, 2015.

\bibitem[Todorov et~al.(2012)Todorov, Erez, and Tassa]{todorov_MuJoCo:_2012}
Emanuel Todorov, Tom Erez, and Yuval Tassa.
\newblock {MuJoCo}: {A} physics engine for model-based control.
\newblock In \emph{2012 {IEEE}/{RSJ} {International} {Conference} on
  {Intelligent} {Robots} and {Systems}}, pp.\  5026--5033, Vilamoura-Algarve,
  Portugal, October 2012. IEEE.
\newblock ISBN 978-1-4673-1736-8 978-1-4673-1737-5 978-1-4673-1735-1.
\newblock \doi{10.1109/IROS.2012.6386109}.
\newblock URL \url{http://ieeexplore.ieee.org/document/6386109/}.

\end{thebibliography}
\bibliographystyle{iclr2020_conference}

\appendix

\section{Assumptions and Definitions}
\label{assumptions_and_definitions}
Reinforcement Learning algorithms vary widely in design, and our metrics are based on certain notions that should span the gamut of RL algorithms. 

\begin{description}
\item[Policy] A policy $\pi_\Theta(a_i|s_i)$ is a distribution over actions $a_i$ given a current (input) state $s_i$. We assume policies are parameterized by a parameter $\Theta$.
\item[Agent] An agent is defined as a distribution over policies (or equivalently a distribution over parameters $\Theta$). In many cases, an agent will be a single policy but for population-based RL methods, the agent is a discrete set of policies. 
\item[Window] A window is a collection of states over which the agent is assumed to have small variation. A window could be a sequence of consecutive time steps for a sequential RL algorithm, or a collection of states at the same training step of a distributed RL algorithm with a parameter server (all agents share $\Theta$).
\item[Performance] The performance of an agent is the mean or median per-epoch reward from running that agent. If the agent is a single policy, then the performance $p(\pi_\Theta)$ is the mean or median per-epoch reward for that agent. 
If the agent is a distribution $D(\Theta)$ of policies, then the performance is the median of $p(\pi_\Theta)$ with $\Theta \sim D$.
\item[Training Run] A training run is a sequence of updates to the agent $D(\Theta)$ from running a reinforcement learning algorithm. It leads to a trained agent $D_{final}(\Theta)$. Multiple training runs share no information with each other. 
\end{description}

We cannot directly measure performance since it is a statistic across an infinite sample of evaluation runs of an agent. Instead we use windows to compute sample medians to approximate performance.

\section{Detrending by differencing}
\label{detrending_by_differencing}

Typically, de-trending can be performed in two main ways \citep{nelson_trends_1982,hamilton_time_1994}. Differencing (i.e. $y_t\prime = y_t - y_{t-1}$) is more appropriate for difference-stationary (DS) processes (e.g. a random walk: $y_t = y_{t-1} + b + \epsilon_t$), where the shocks $\epsilon_t$ accumulate over time. For trend-stationary (TS) processes, which are characterized by stationary fluctuations around a deterministic trend, e.g. $y_t = a + b*t + \epsilon_t$, it is more appropriate to fit and subtract that trend. 

We performed an analysis of real training runs and verified that the data are indeed approximately DS, and that differencing does indeed remove the majority of time-dependent structure. For this analysis we used the training runs on Atari as described in \ref{atari_data}. Before differencing, the Augmented Dickey-Fuller test (ADF test, also known as a difference-stationarity test; \citet{said_e._said_testing_1984}) rejects the null hypothesis of a unit root on only 72\% of the runs; after differencing, the ADF test rejects the null hypothesis on 92\% of the runs (p-value threshold 0.05). For the ADF test, the rejection of a unit root (of the autoregressive lag polynomial) implies the alternate hypothesis, which is that the time series is trend-stationary.

Therefore, our training curves are better characterized as an accumulation of shocks, i.e. as DS processes, rather than as mean-reverting TS processes. They are not actually purely DS because the shocks $\epsilon_t$ are not stationary over time, but because we compute standard deviation within sliding windows, we can capture the non-stationarity and change in variability over time. Thus, we chose to detrend using differencing.

As a further note in favor of detrending by differencing, it is useful to observe that many measures of variability are defined relative to the central tendency of the data, e.g. the median absolute deviation $\text{MAD} = \text{median}(|X_i - \widetilde{X}|)$ where $\widetilde{X}$ is the median of $X$. On the raw data (without differencing), the MAD would be defined relative to $\widetilde{X}$ as median performance, so that any improvements in performance are included in that computation of variability. On the other hand, if we compute MAD on the 1st-order differences, we are using a $\widetilde{X}$ that represents the median \emph{change} in performance, which is a more reasonable baseline to compute variability against, when we are in fact concerned with the variability of those changes. 

A final benefit of differencing is that it is parameter-free.

\section{Illustrations of Permutation Test Procedures}
\label{permutation_test_illustrations}

\begin{figure}[htp]
    \centering
    \begin{subfigure}[t]{\textwidth}
        \includegraphics[width=\textwidth]{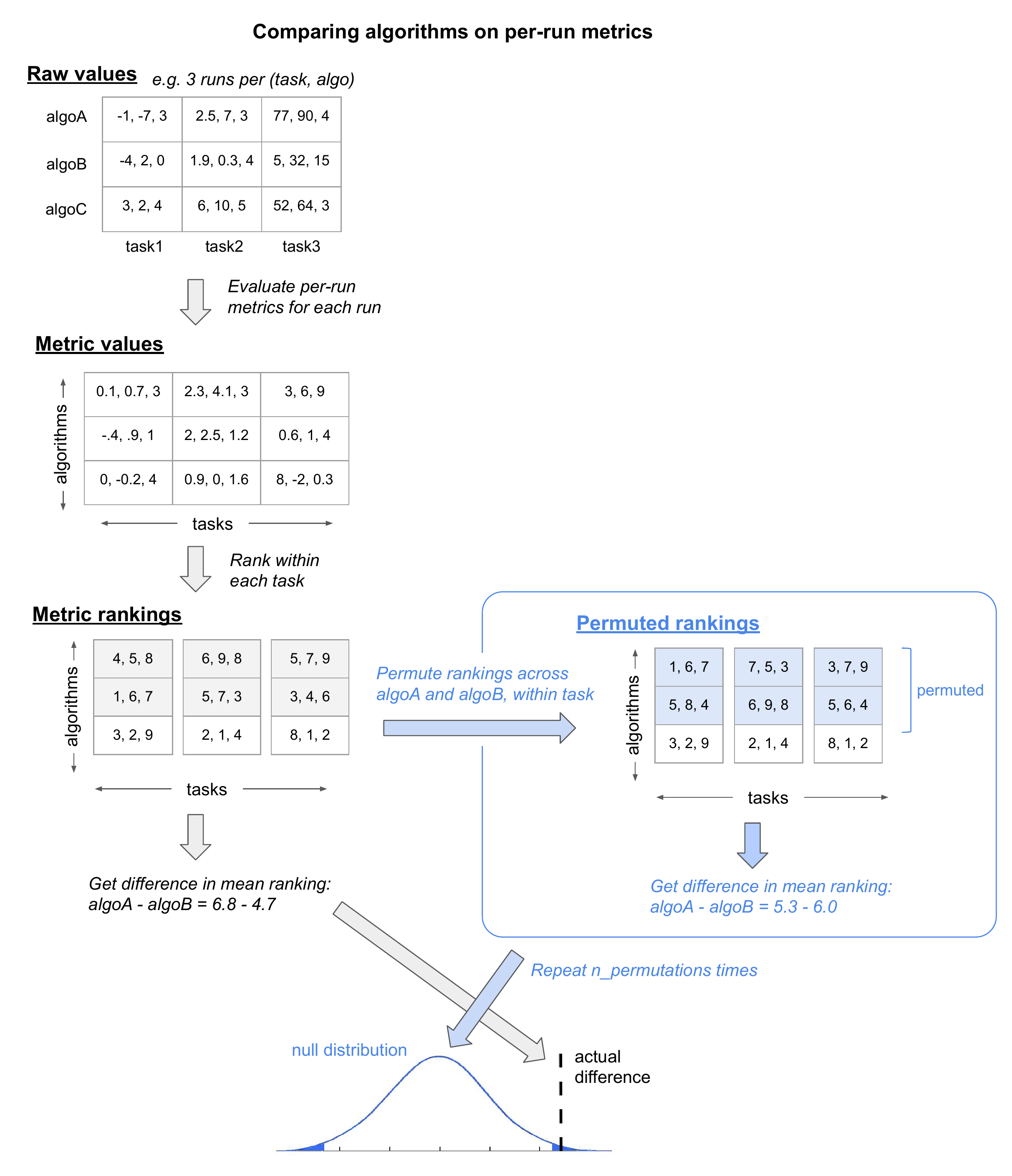}
    \end{subfigure}
    \caption{Diagram illustrating the computation of the permutation tests for \textbf{per-run metrics} (Dispersion across Time, Short-term Risk across Time, Long-term Risk across Time). In this example, we are comparing Algorithm A and Algorithm B, and there are only 3 algorithms, 3 tasks, and 3 runs per (task, algo) pair. To compute the difference in average rankings for two algorithms, follow the gray arrows. To compute a null distribution of difference in average rankings (by permuting the runs), follow the blue arrows a number of times (e.g. 1,000 times). Once the null distribution has been computed, the actual value of the difference can be compared with the null distribution to obtain a p-value.}
    \label{fig_comparing_per_run_metrics}
\end{figure}

\begin{figure}[htp]
    \centering
    \begin{subfigure}[t]{\textwidth}
        \includegraphics[width=\textwidth]{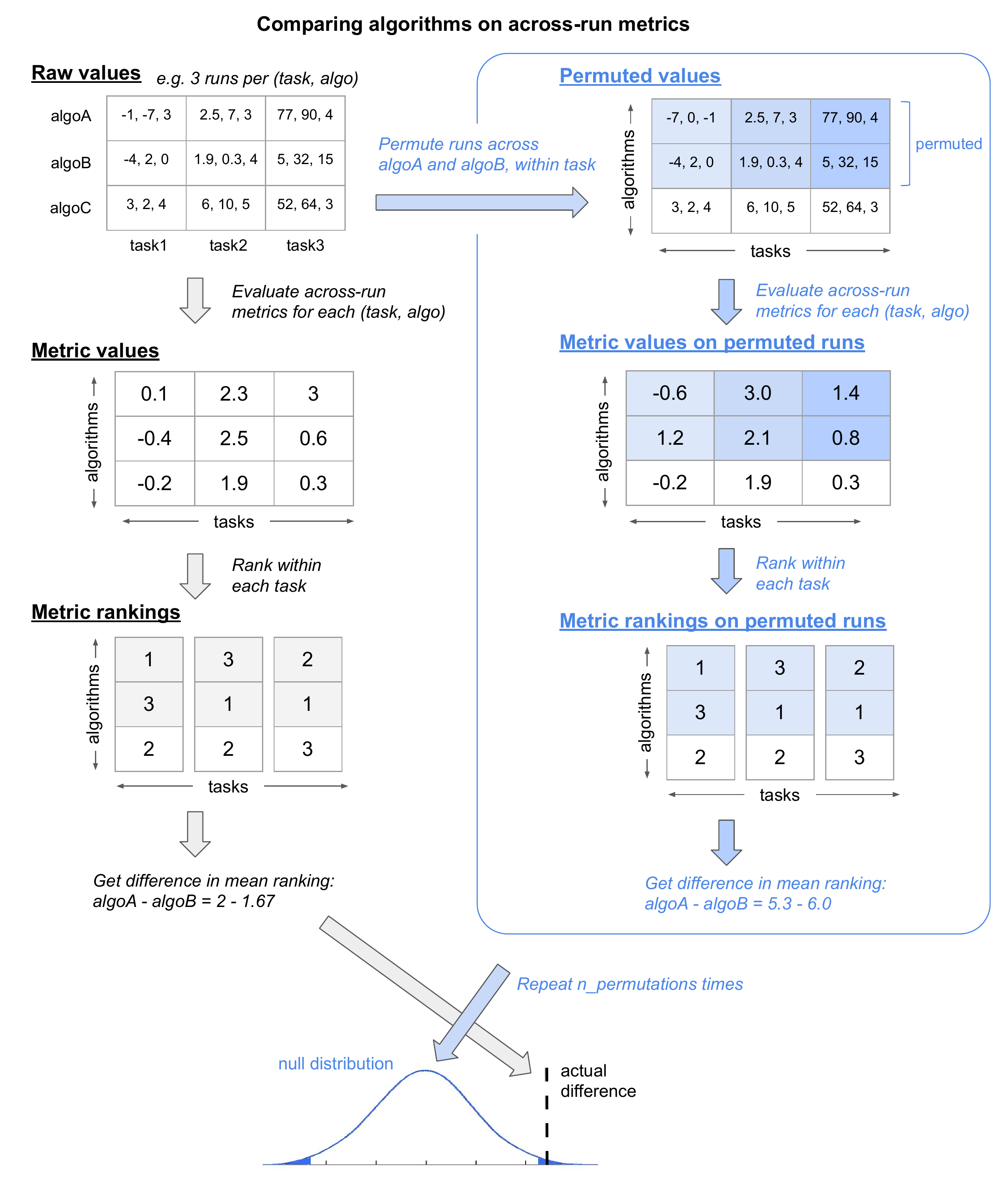}
    \end{subfigure}
    \caption{Diagram illustrating the computation of the permutation tests for \textbf{across-run or across-rollout metrics} (Dispersion across Runs, Risk Across Runs, Dispersion across Fixed-policy rollouts, Risk across Fixed-Policy rollouts). In this example, we are comparing Algorithm A and Algorithm B, and there are only 3 algorithms, 3 tasks, and 3 runs per (task, algo) pair. To compute the difference in average rankings for two algorithms, follow the gray arrows. To compute a null distribution of difference in average rankings (by permuting the runs), follow the blue arrows a number of times (e.g. 1,000 times). Once the null distribution has been computed, the actual value of the difference can be compared with the null distribution to obtain a p-value.}
    \label{fig_comparing_across_run_metrics}
\end{figure}

We illustrate the procedure for computing permutation tests to compare pairs of algorithms on a specified metric, in Figs. \ref{fig_comparing_per_run_metrics} (for per-run metrics) and \ref{fig_comparing_across_run_metrics} (for across-run metrics).

\section{Raw training curves for OpenAI MuJoCo tasks}
\label{raw_mujoco_curves}

\begin{figure}[htp!]
    \includegraphics[width=\textwidth]{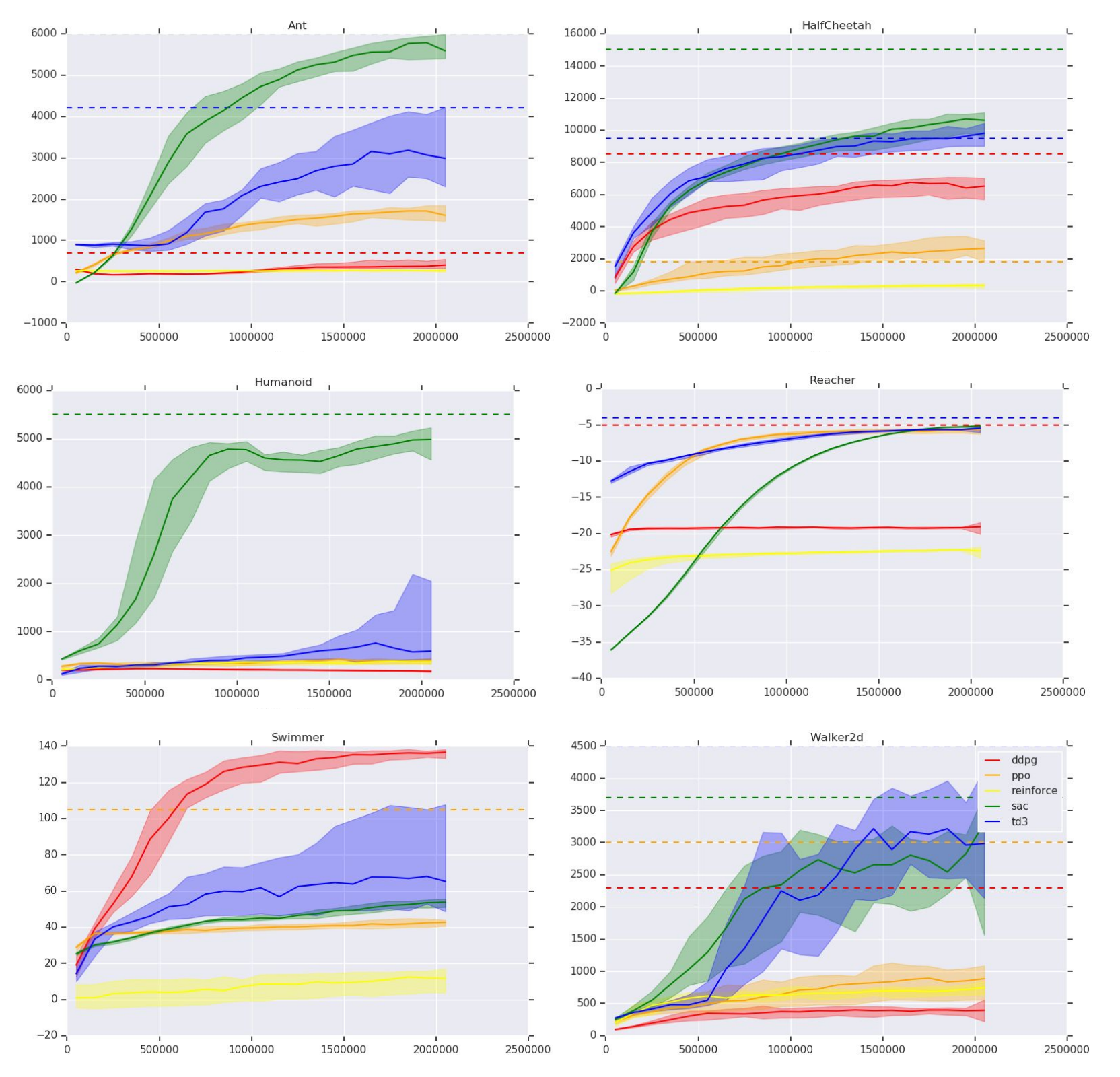}
    \caption{Raw training curves for OpenAI MuJoCo tasks. The x-axes indicate environment steps, and the y-axes indicate average per-episode return. Dotted lines indicate baseline performance from the literature, where available.}
    \label{fig_raw_mujoco_curves}
\end{figure}

In Figure \ref{fig_raw_mujoco_curves}, we show the raw training curves for the TF-Agents implementations of continuous-control algorithms, applied to the OpenAI MuJoCo tasks. These are compared against baselines from the literature, where available (DDPG and TD3: \citet{fujimoto_addressing_2018}, PPO: \citet{schulman_proximal_2017}, SAC: \citet{haarnoja_soft_2018})

\section{Hyperparameter settings}
\label{hyperparameters}
% \label{vizier_trials}

% The black-box optimization of hyperparameters (on continuous control algorithms applied to OpenAI Gym tasks, described in Section \ref{MuJoCo_data}) provided us with a set of training runs each trained with a different set of hyperparameters. This allowed us to apply the same reliability metrics to explore reliability across and sensitivity to hyperparameters. We measured dispersion (IQR) and risk (CVaR) across hyperparameters, and the results are shown in Figure \ref{fig_vizier_trials}. Again, the best performing algorithms by median performance do not have the best reliability across hyperparameters. 

% Note that the distribution of hyperparameters in this analysis is highly dependent on the details of the black-box optimization. However, we argue that this distribution of hyperparameters is an approximation to hyperparameter settings that may be chosen by other black-box optimizers, and also to hyperparameters chosen manually. Note that this analysis is also highly dependent upon the configuration of the search space. Thus, these results should be interpreted as measuring reliability of these algorithms \textit{given this particular search space}. The hyperparameter search space is shown in Table \ref{vizier_search_space}.

For the continuous control experiments, hyperparameters were chosen on a per-environment basis according to the black-box optimization algorithm described in \citet{golovin_google_2017}. The hyperparameter search space is shown in Table \ref{vizier_search_space}.

For the discrete control experiments, hyperparameter selection is described in \citep{dopamine}. Hyperparameters are shown in Table \ref{dopamine_hyperparams}, duplicated for reference from https://github.com/google/dopamine/tree/master/baselines.

% \begin{figure}[ht]
%     \centering
%     \begin{subfigure}[t]{0.84\textwidth}
%         \includegraphics[width=\textwidth]{figures/mujoco_vizier.pdf}
%     \end{subfigure}
%     \caption{Reliability metrics and median performance across hyperparameter settings, for continuous control RL algorithms tested on OpenAI Gym environments. Hyperparameters were chosen according to the black-box optimization algorithm described in \citet{golovin_google_2017}. Rank 1 always indicates "best" reliability. Error bars are 95\% bootstrap confidence intervals (\# bootstraps = 1,000). Significant pairwise differences in ranking between pairs of algorithms are indicated by black horizontal lines above the colored bars. ($\alpha=0.05$ with Benjamini-Yekutieli correction, permutation test with \# permutations = 1,000).}
%     \label{fig_vizier_trials}
% \end{figure}

\begin{table}[bp]
\caption{Hyperparameter search space for continuous control algorithms.}
\label{vizier_search_space}
\begin{center}
\begin{tabular}{|c|c|c|c|c|}
\hline
\textbf{Algorithm}       &\textbf{Hyperparameter}             &\textbf{Search min} & \textbf{Search max}\\
\hline
SAC             &actor learning rate        &0.000001   &0.001  \\
                &$\alpha$ learning rate        &0.000001   &0.001  \\
                &critic learning rate       &0.000001   &0.001  \\
                &target update $\tau$          &0.00001    &1.0  \\
\hline
TD3             &actor learning rate        &0.000001   &0.001  \\
                &critic learning rate       &0.000001   &0.001  \\
                &target update $\tau$          &0.00001    &1.0  \\
\hline
PPO             &learning rate         &0.000001       &0.001  \\
\hline
DDPG            &actor learning rate        &0.000001   &0.001  \\
                &critic learning rate       &0.000001   &0.001  \\
                &target update $\tau$          &0.00001    &1.0  \\
\hline
REINFORCE        &learning rate         &0.000001       &0.001  \\
                 &\# episodes before each train step  &1.0      &10  \\
\hline
\end{tabular}
\end{center}
\end{table}

\begin{table}[htbp]
\caption{Final hyperparameters for SAC.}
\label{sac_mujoco_hyperparams}
\begin{center}
\begin{tabular}{|c|c|c|c|c|}
\hline
                &actor learning rate    &$\alpha$ learning rate    &critic learning rate   &target update $\tau$ \\
\hline
Ant-v2          &0.000006               &0.000009               &0.0009                 &0.0002 \\
HalfCheetah-v2  &0.0001                 &0.000005               &0.0004                 &0.02 \\
Humanoid-v2     &0.0003                 &0.0008                 &0.0006                 &0.8 \\
Reacher-v2      &0.00001                &0.000002               &0.0005                 &0.00002 \\
Swimmer-v2      &0.000004               &0.000009               &0.0002                 &0.009 \\
Walker2d-v2     &0.0002                 &0.0009                 &0.0008                 &0.01 \\
\hline
\end{tabular}
\end{center}
\end{table}

\begin{table}[htbp]
\caption{Final hyperparameters for TD3.}
\label{td3_mujoco_hyperparams}
\begin{center}
\begin{tabular}{|c|c|c|c|}
\hline
                &actor learning rate    &critic learning rate   &target update $\tau$ \\
\hline
Ant-v2          &0.000001               &0.0002                 &0.0003 \\
HalfCheetah-v2  &0.0003                 &0.0005                 &0.02 \\
Humanoid-v2     &0.0001                 &0.0001                 &0.0002 \\
Reacher-v2      &0.000001               &0.00003                &0.00003 \\
Swimmer-v2      &0.0004                 &0.0002                 &0.01 \\
Walker2d-v2     &0.00006                &0.00009                &0.001 \\
\hline
\end{tabular}
\end{center}
\end{table}

\begin{table}[htbp]
\caption{Final hyperparameters for PPO.}
\label{ppo_mujoco_hyperparams}
\begin{center}
\begin{tabular}{|c|c|c|c|c|c|c|}
\hline
                &learning rate \\
\hline
Ant-v2          &0.0008 \\
HalfCheetah-v2  &0.0008 \\
Humanoid-v2     &0.0008 \\
Reacher-v2      &0.00002 \\
Swimmer-v2      &0.0004 \\
Walker2d-v2     &0.0002 \\
\hline
\end{tabular}
\end{center}
\end{table}

\clearpage
\begin{table}[htbp]
\caption{Final hyperparameters for DDPG.}
\label{ddpg_mujoco_hyperparams}
\begin{center}
\begin{tabular}{|c|c|c|c|}
\hline
                &actor learning rate    &critic learning rate   &target update $\tau$ \\
\hline
Ant-v2          &0.00003                &0.0004                 &0.0002 \\
HalfCheetah-v2  &0.00006                &0.0005                 &0.02 \\
Humanoid-v2     &0.00006                &0.00009                &0.01 \\
Reacher-v2      &0.00005                &0.0005                 &0.005 \\
Swimmer-v2      &0.0005                 &0.0003                 &0.004 \\
Walker2d-v2     &0.0003                 &0.0004                 &0.03 \\
\hline
\end{tabular}
\end{center}
\end{table}

\begin{table}[htbp]
\caption{Final hyperparameters for REINFORCE.}
\label{reinforce_mujoco_hyperparams}
\begin{center}
\begin{tabular}{|c|c|c|c|}
\hline
                &learning rate  &\# episodes before \\
                &               &each train step \\
\hline
Ant-v2          &0.00002                 &9  \\
HalfCheetah-v2  &0.0004                 &7 \\
Humanoid-v2     &0.0005                &2 \\
Reacher-v2      &0.000004                 &6 \\
Swimmer-v2      &0.000005                 &3 \\
Walker2d-v2     &0.0001                 &6 \\
\hline
\end{tabular}
\end{center}
\end{table}

\begin{table}[!htb]
\caption{Hyperparameters for discrete control algorithms.}
\label{dopamine_hyperparams}
\begin{center}
\begin{tabular}{|c|c|c|c|c|}
\hline
Training $\epsilon$     &Evaluation $\epsilon$      &$\epsilon$ decay schedule      &Min. history       &Target network  \\
                        &                           &                               &to start learning  &update frequency\\
\hline
0.01                    &0.001               &1,000,000 frames               &80,000 frames                 &32,000 frames \\
\hline
\end{tabular}
\end{center}
\end{table}

\vspace{20mm}

\section{Per-task metric results}
\label{per_task_results}

\begin{figure}[htbp]
    \centering
    \begin{subfigure}[t]{0.95\textwidth}
        \centering
        \includegraphics[width=\textwidth]{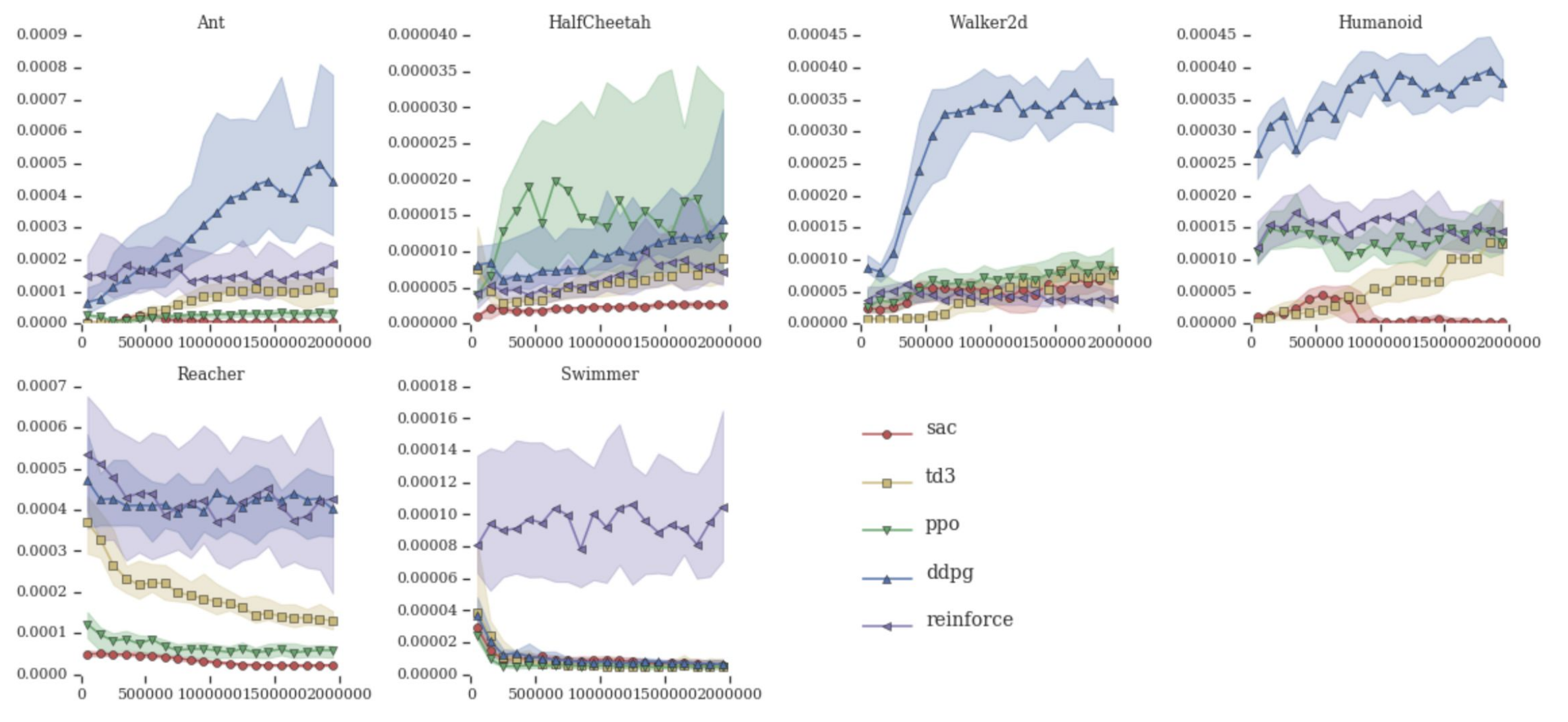}
        \caption{Dispersion across Time. Better reliability is indicated by less positive values. The x-axes indicate the number of environment steps.}
    \end{subfigure}
    \begin{subfigure}[t]{0.95\textwidth}
        \includegraphics[width=\textwidth]{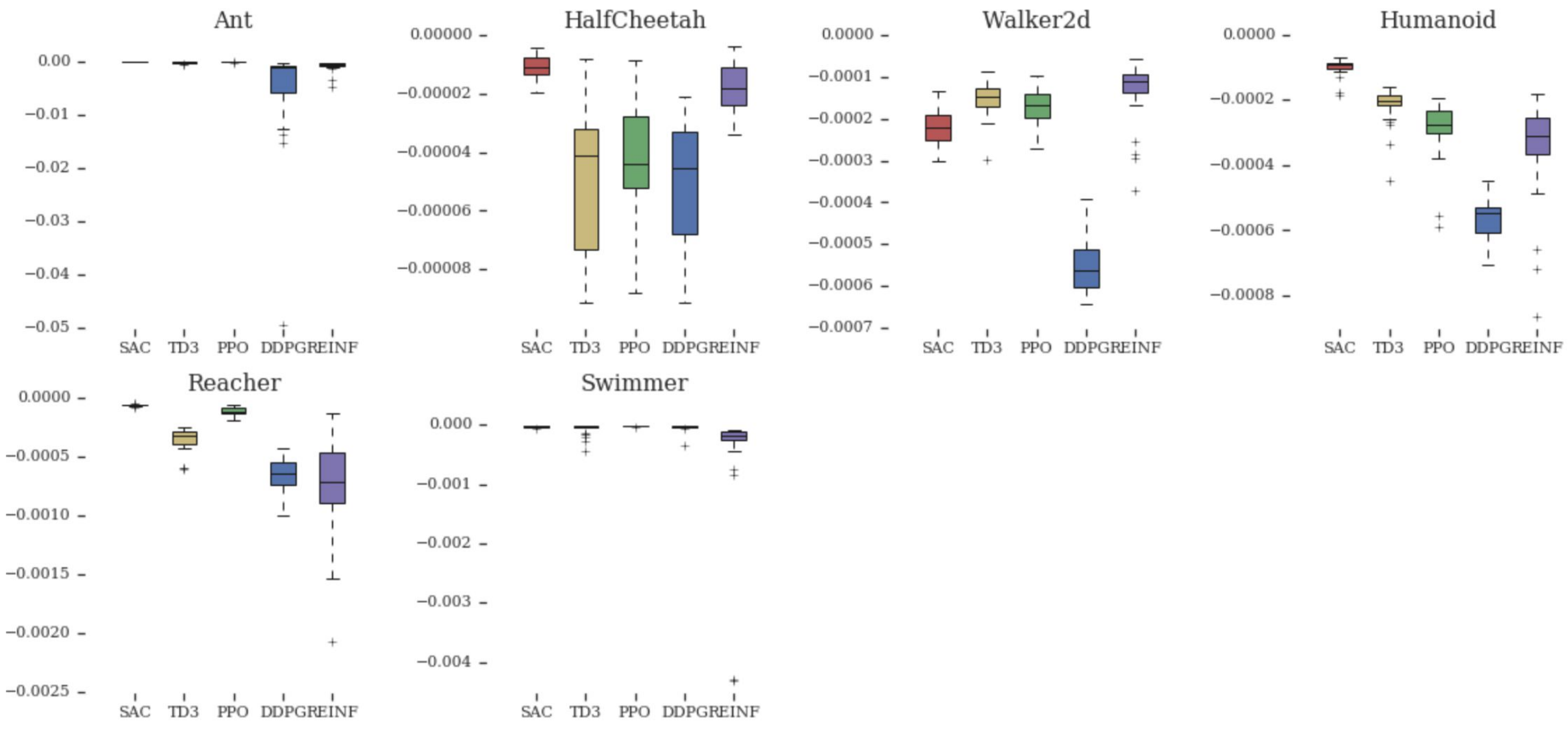}
        \caption{Short-term Risk across Time. Better reliability is indicated by more positive values.}
    \end{subfigure}
    \begin{subfigure}[t]{0.95\textwidth}
        \includegraphics[width=\textwidth]{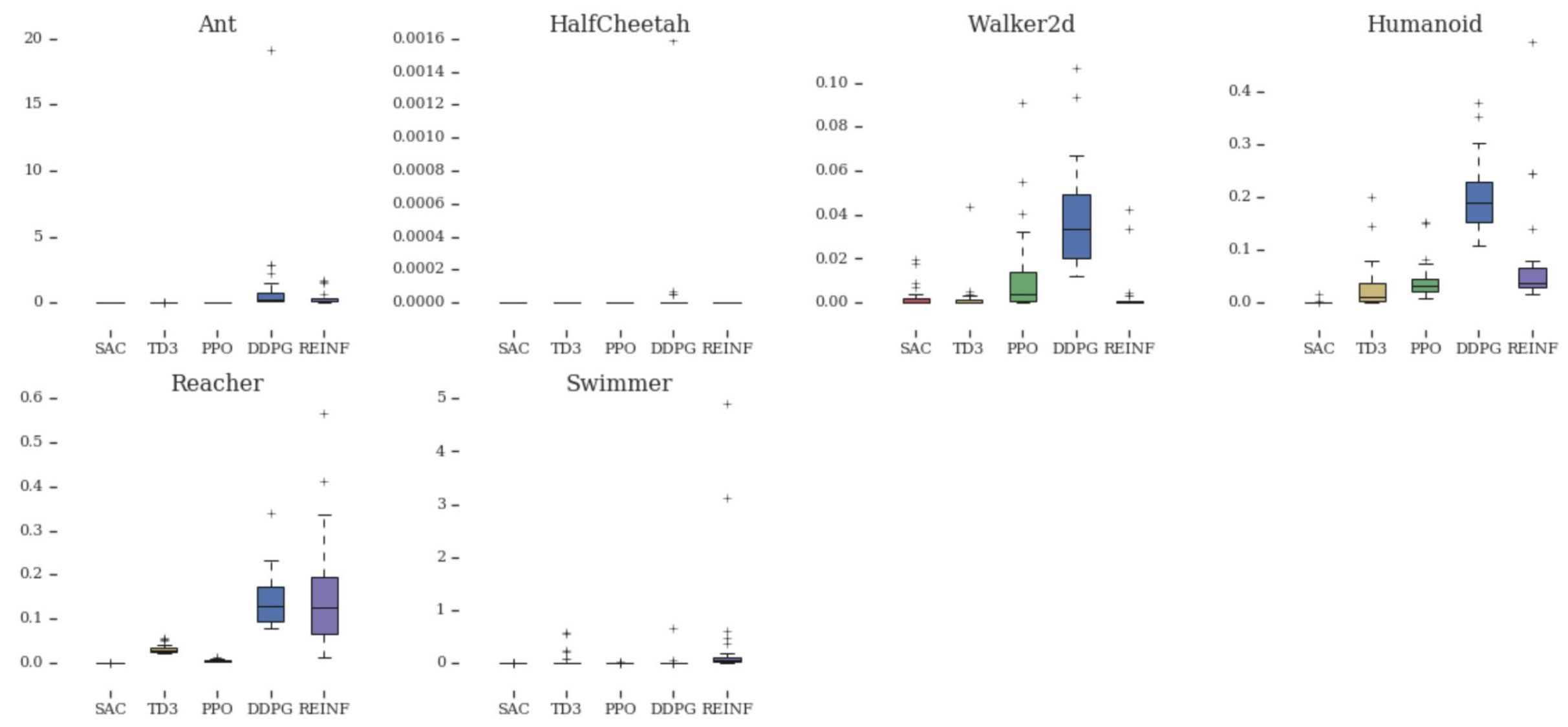}
        \caption{Long-term Risk across Time. Better reliability is indicated by less positive values.}
    \end{subfigure}
    \caption{Across-time reliability metrics for continuous control RL algorithms tested on OpenAI Gym environments, evaluated on a per-environment basis.}
    \label{mujoco_per_task_across_time}
\end{figure}

\begin{figure}[htbp]
    \centering
    \begin{subfigure}[t]{0.95\textwidth}
        \centering
        \includegraphics[width=\textwidth]{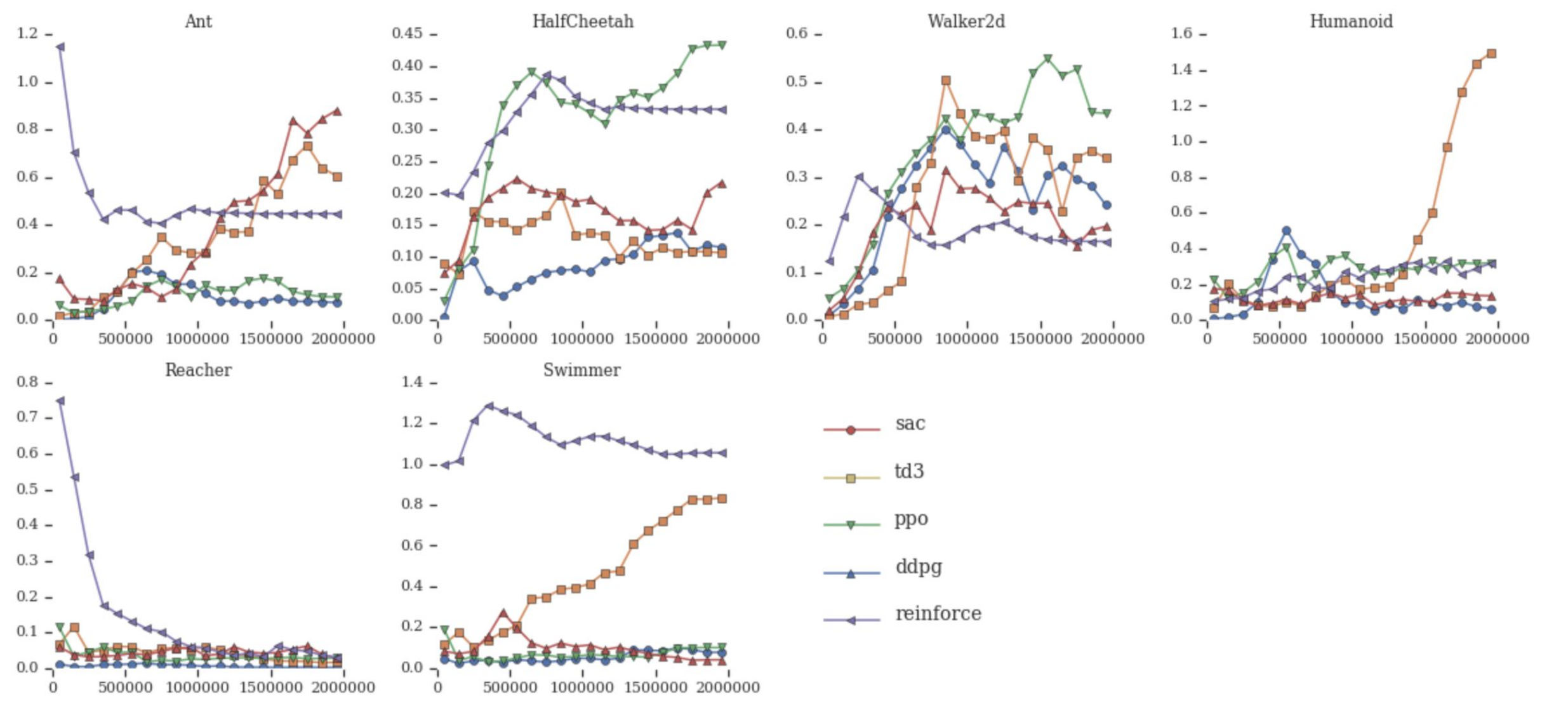}
        \caption{Dispersion across Runs. Better reliability is indicated by less positive values.}
        \label{mujoco_per_task_DR}
    \end{subfigure}
    \begin{subfigure}[t]{0.95\textwidth}
        \includegraphics[width=\textwidth]{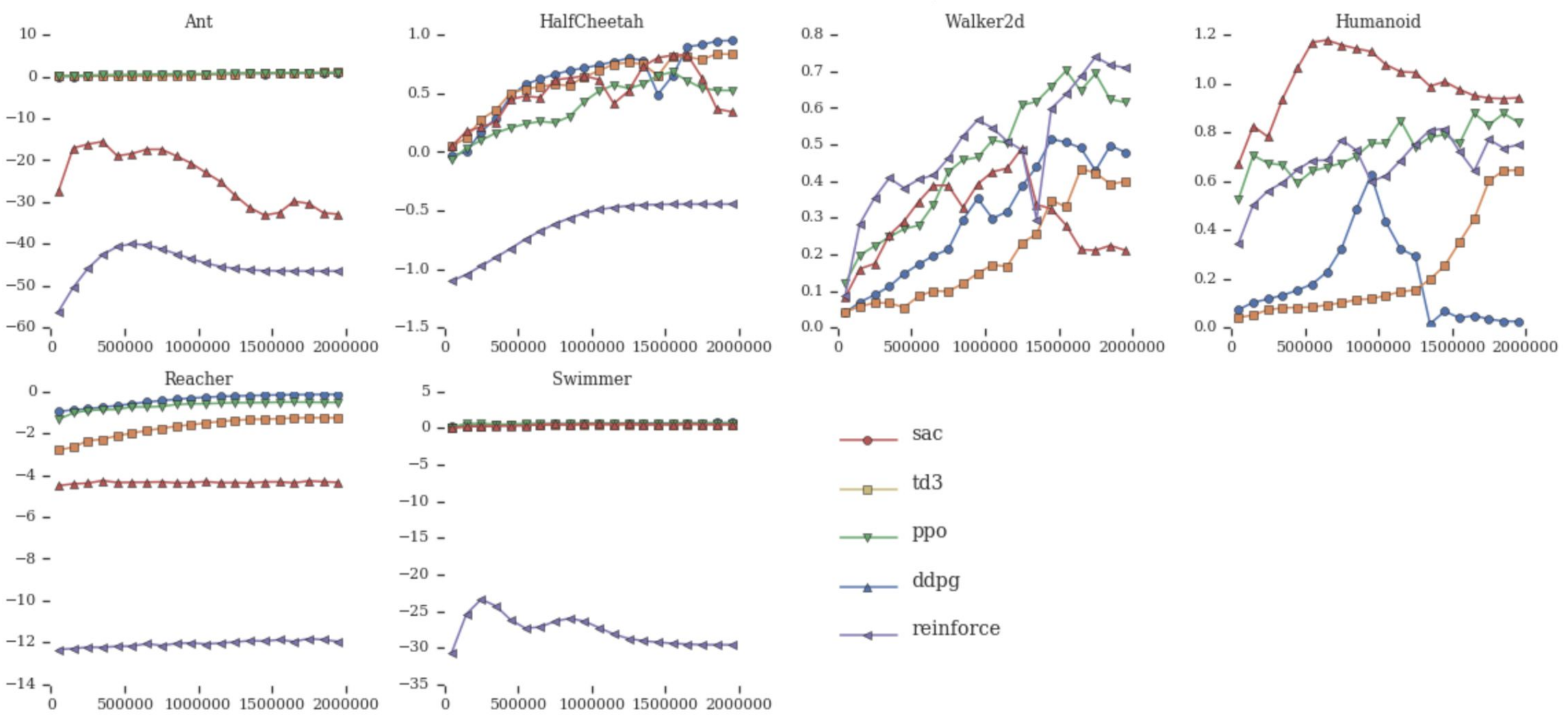}
        \caption{Risk across Runs. Better reliability is indicated by more positive values.}
    \end{subfigure}
    \begin{subfigure}[t]{0.95\textwidth}
        \includegraphics[width=\textwidth]{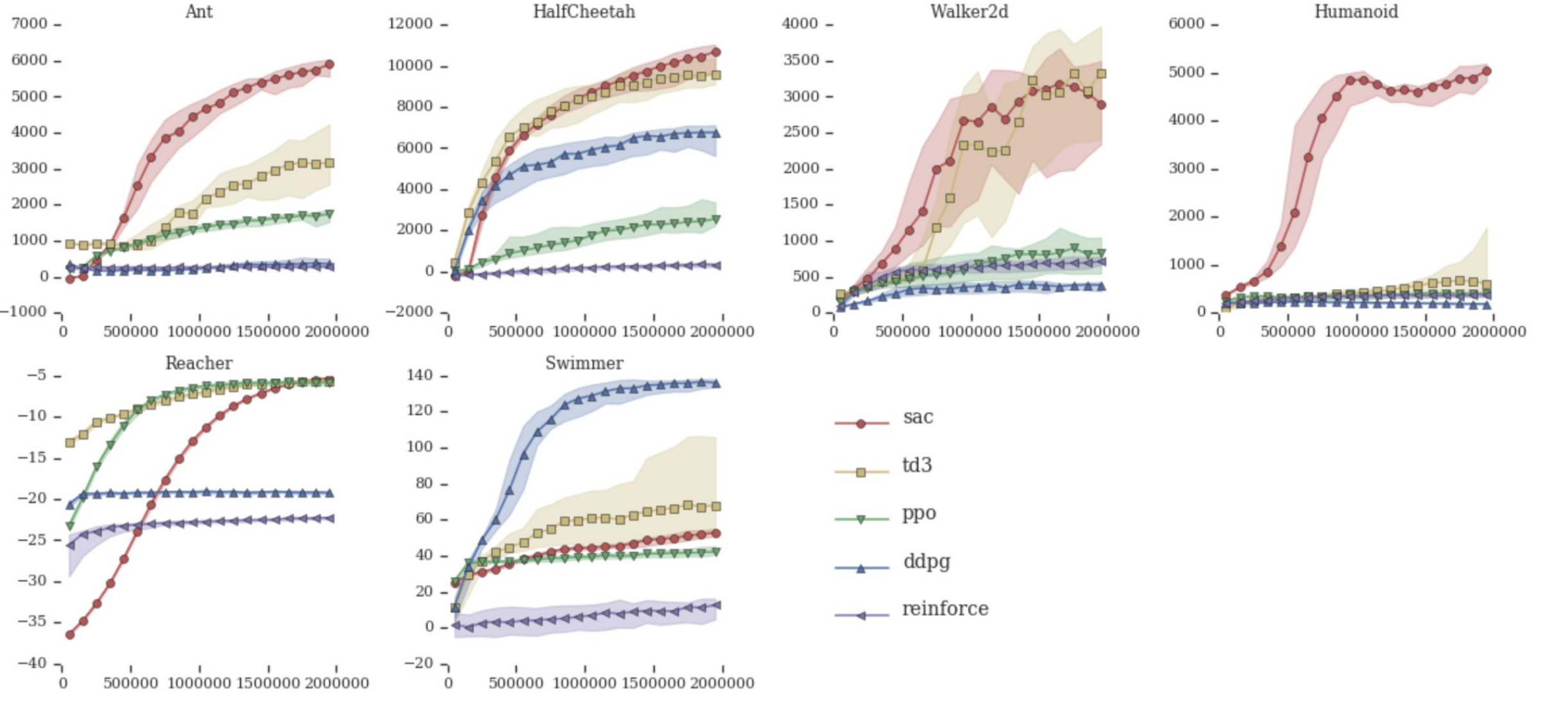}
        \caption{Median performance during training. Better performance is indicated by more positive values.}
    \end{subfigure}
    \caption{Across-run reliability metrics and median performance for continuous control RL algorithms tested on OpenAI Gym environments, evaluated on a per-environment basis. The x-axes indicate the number of environment steps.}
    \label{mujoco_per_task_across_runs}
\end{figure}

\begin{figure}[htbp]
    \centering
    \begin{subfigure}[t]{0.95\textwidth}
        \centering
        \includegraphics[width=\textwidth]{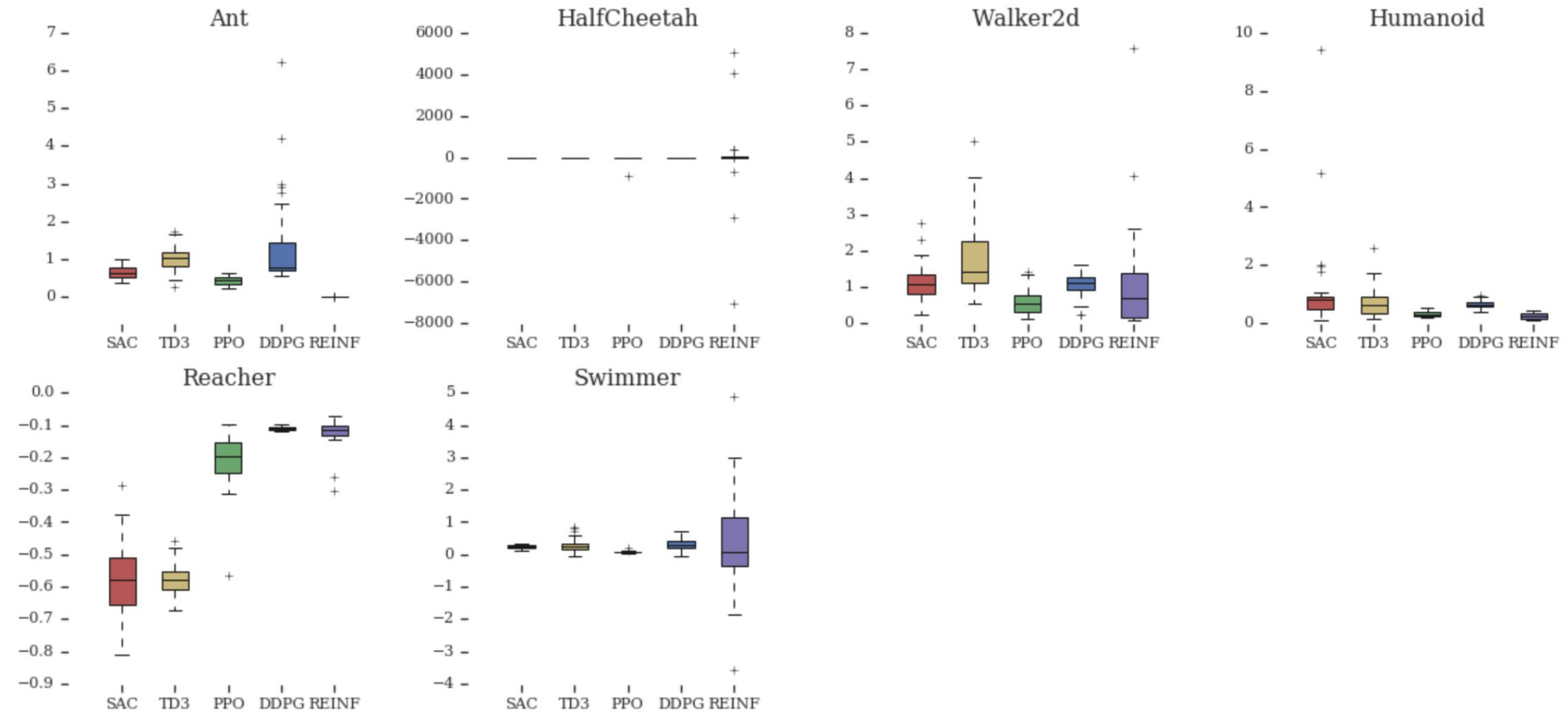}
        \caption{Dispersion on Fixed-policy rollouts. Better reliability is indicated by less positive values.}
    \end{subfigure}
    \begin{subfigure}[t]{0.95\textwidth}
        \includegraphics[width=\textwidth]{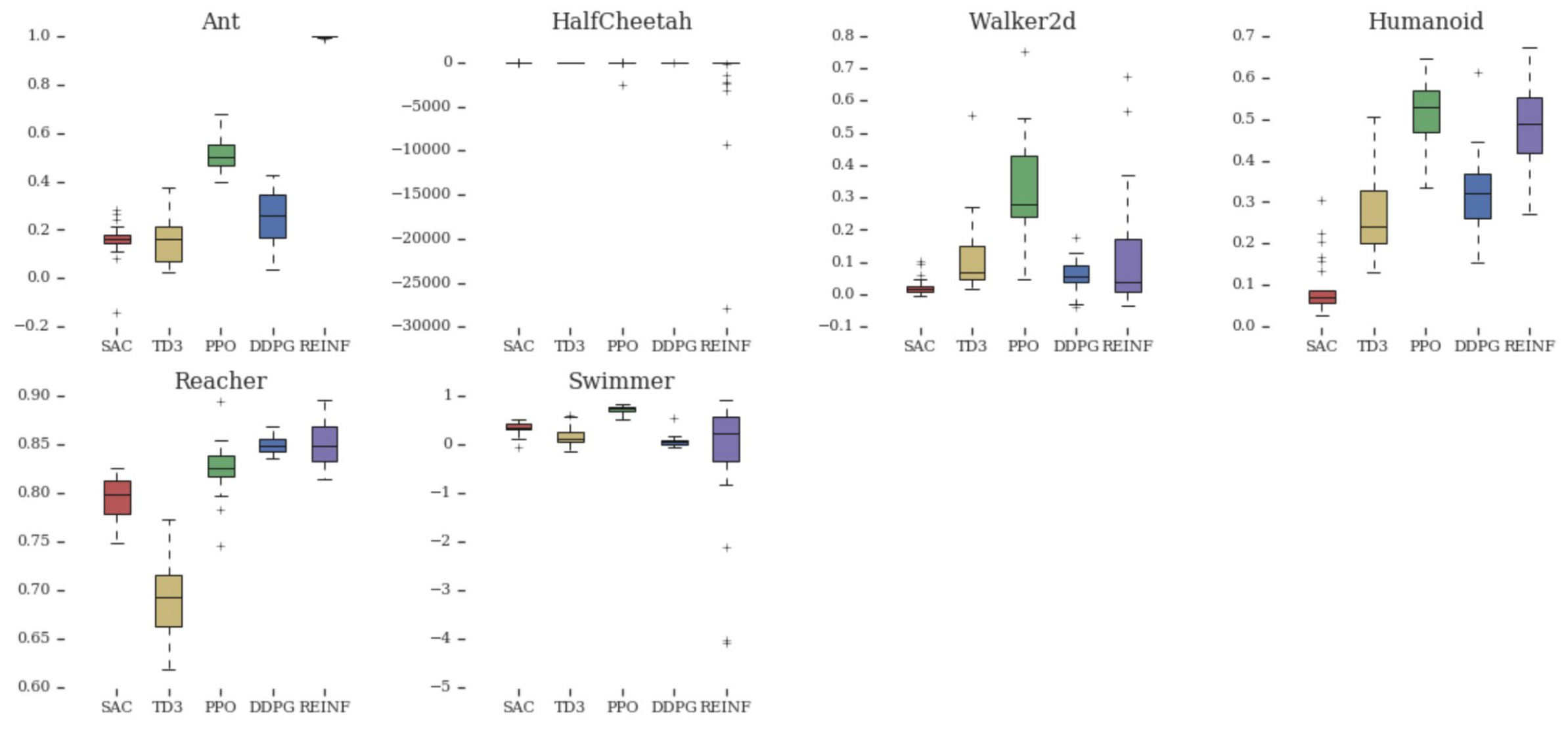}
        \caption{Risk on Fixed-policy rollouts. Better reliability is indicated by more positive values.}
    \end{subfigure}
    \begin{subfigure}[t]{0.95\textwidth}
        \includegraphics[width=\textwidth]{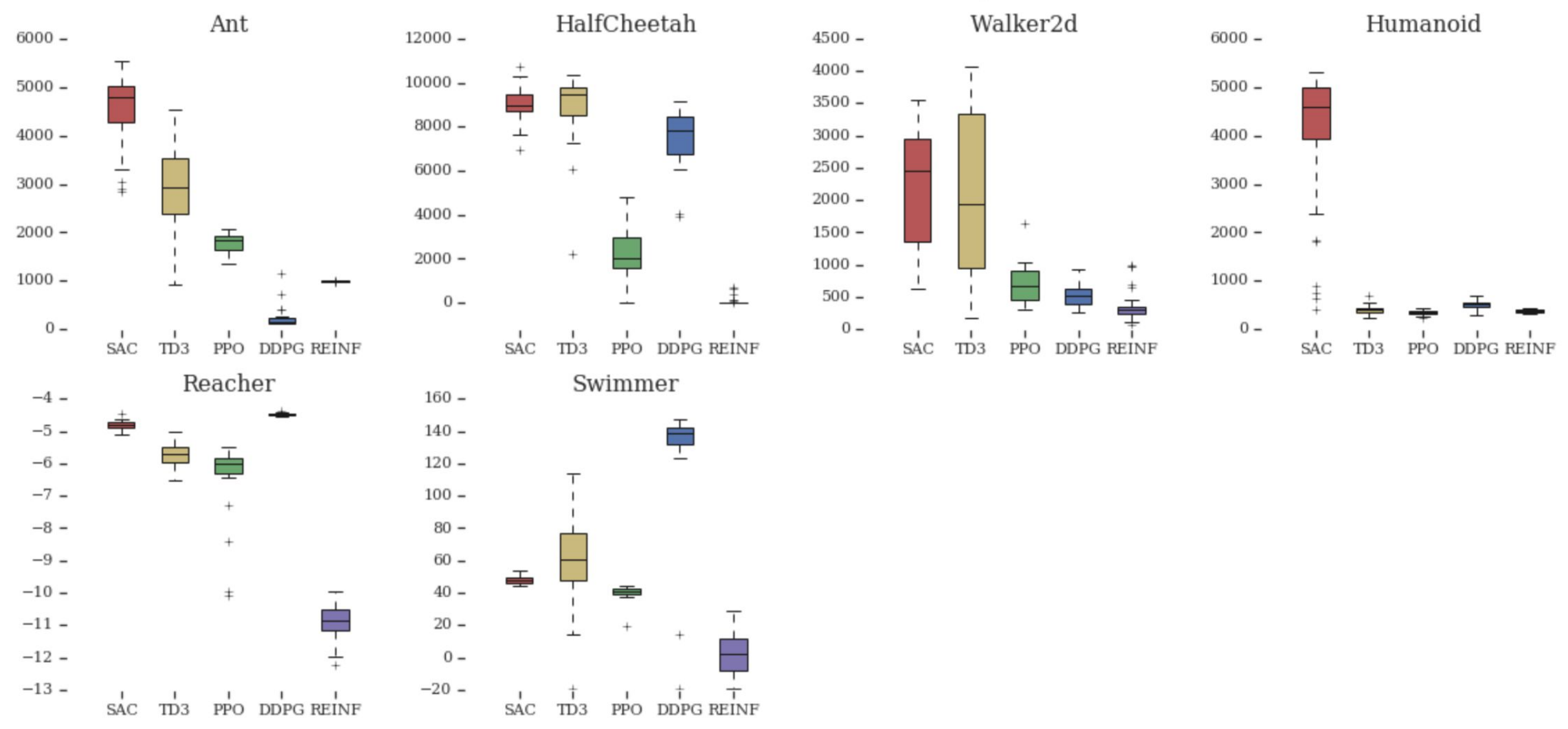}
        \caption{Median performance on Fixed-policy rollouts. Better performance is indicated by more positive values.}
    \end{subfigure}
    \caption{Reliability metrics and median performance on fixed-policy rollouts for continuous control RL algorithms tested on OpenAI Gym environments, evaluated on a per-environment basis.}
    \label{mujoco_per_task_fixed_rollouts}
\end{figure}

\begin{figure}[!ht]
    \centering
    \includegraphics[width=\textwidth]{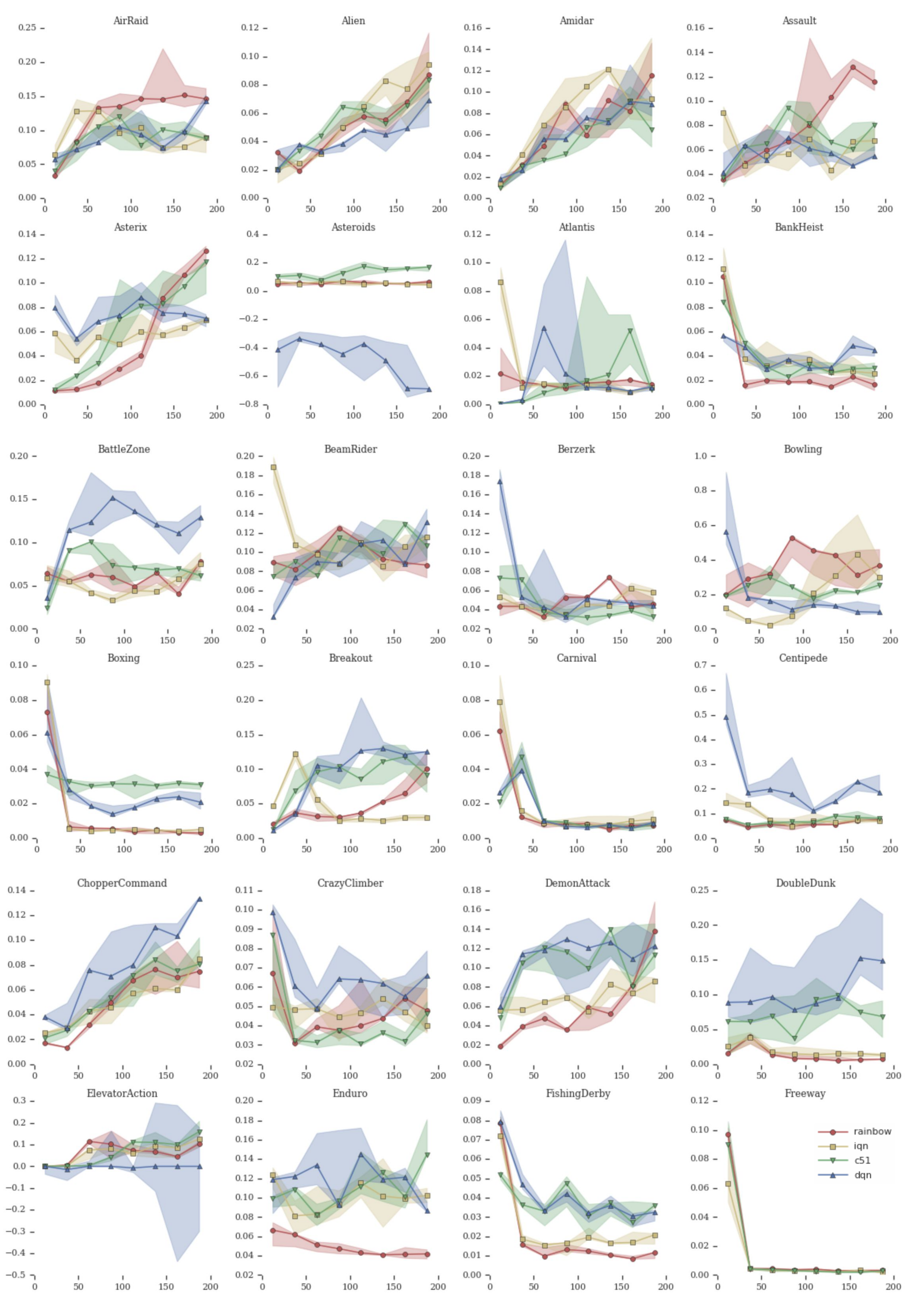}
    \caption{Dispersion across Time for DQN-variants tested on 60 Atari games, evaluated on a per-environment basis (page 1). Better reliability is indicated by less positive values. The x-axes indicate millions of Atari frames.}
    \label{atari_per_task_DT1}
\end{figure}

\begin{figure}[!ht]
    \centering
    \includegraphics[width=\textwidth]{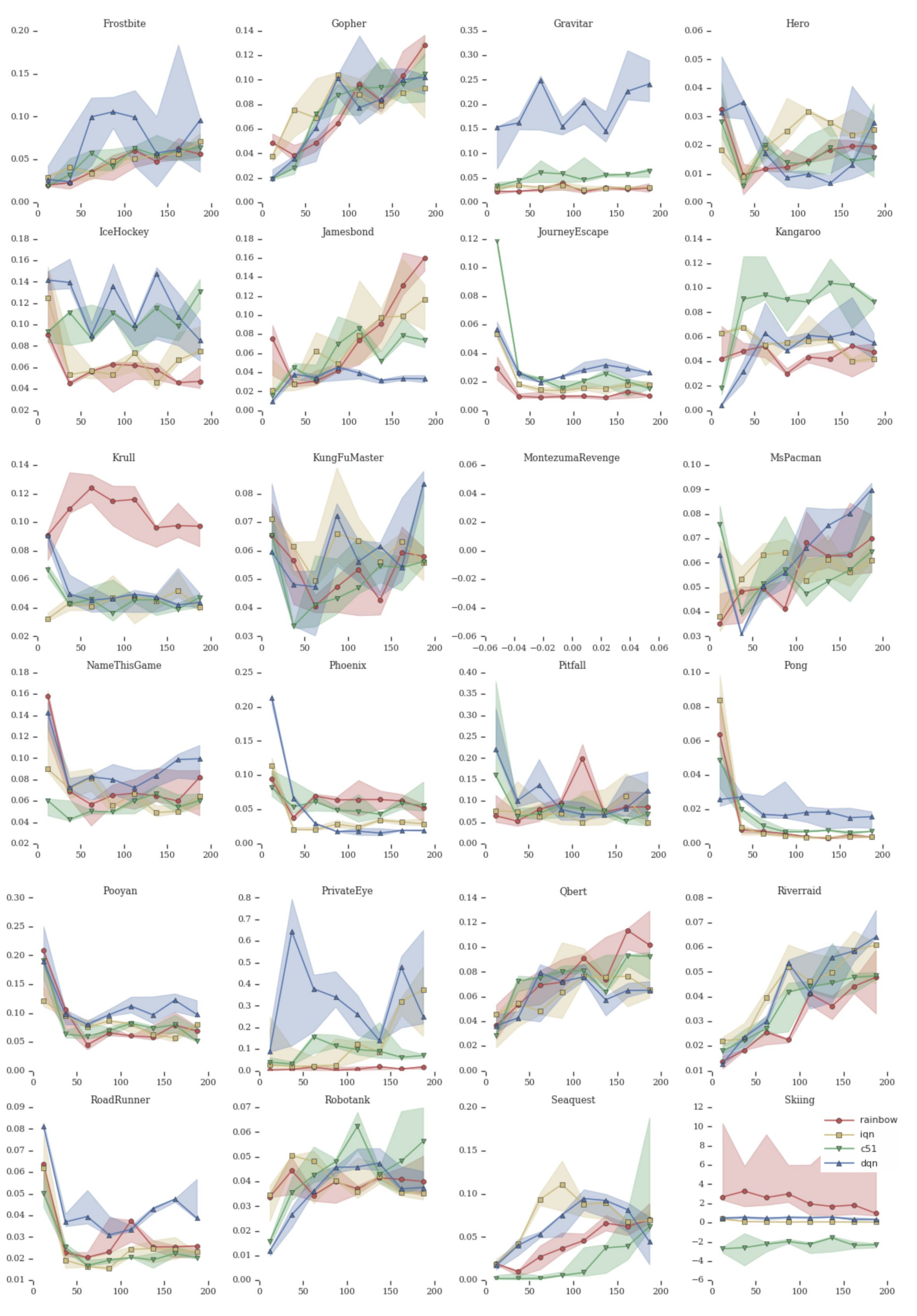}
    \caption{Dispersion across Time for DQN-variants tested on 60 Atari games, evaluated on a per-environment basis (page 2). Better reliability is indicated by less positive values. The x-axes indicate millions of Atari frames.}
    \label{atari_per_task_DT2}
\end{figure}

\begin{figure}[!ht]
    \centering
    \includegraphics[width=\textwidth]{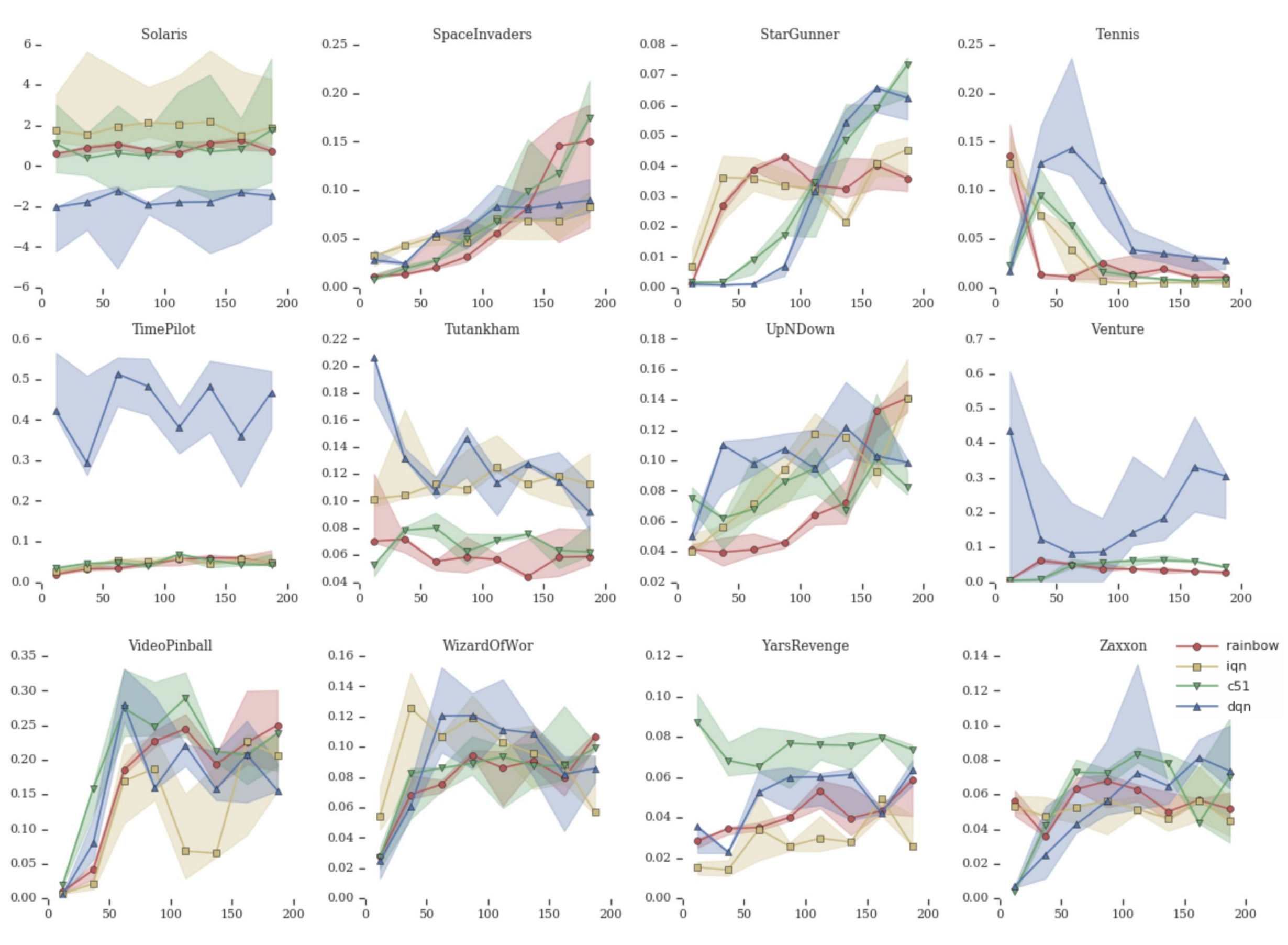}
    \caption{Dispersion across Time for DQN-variants tested on 60 Atari games, evaluated on a per-environment basis (page 3). Better reliability is indicated by less positive values. The x-axes indicate millions of Atari frames.}
    \label{atari_per_task_DT3}
\end{figure}

\begin{figure}[!ht]
    \centering
    \includegraphics[width=\textwidth]{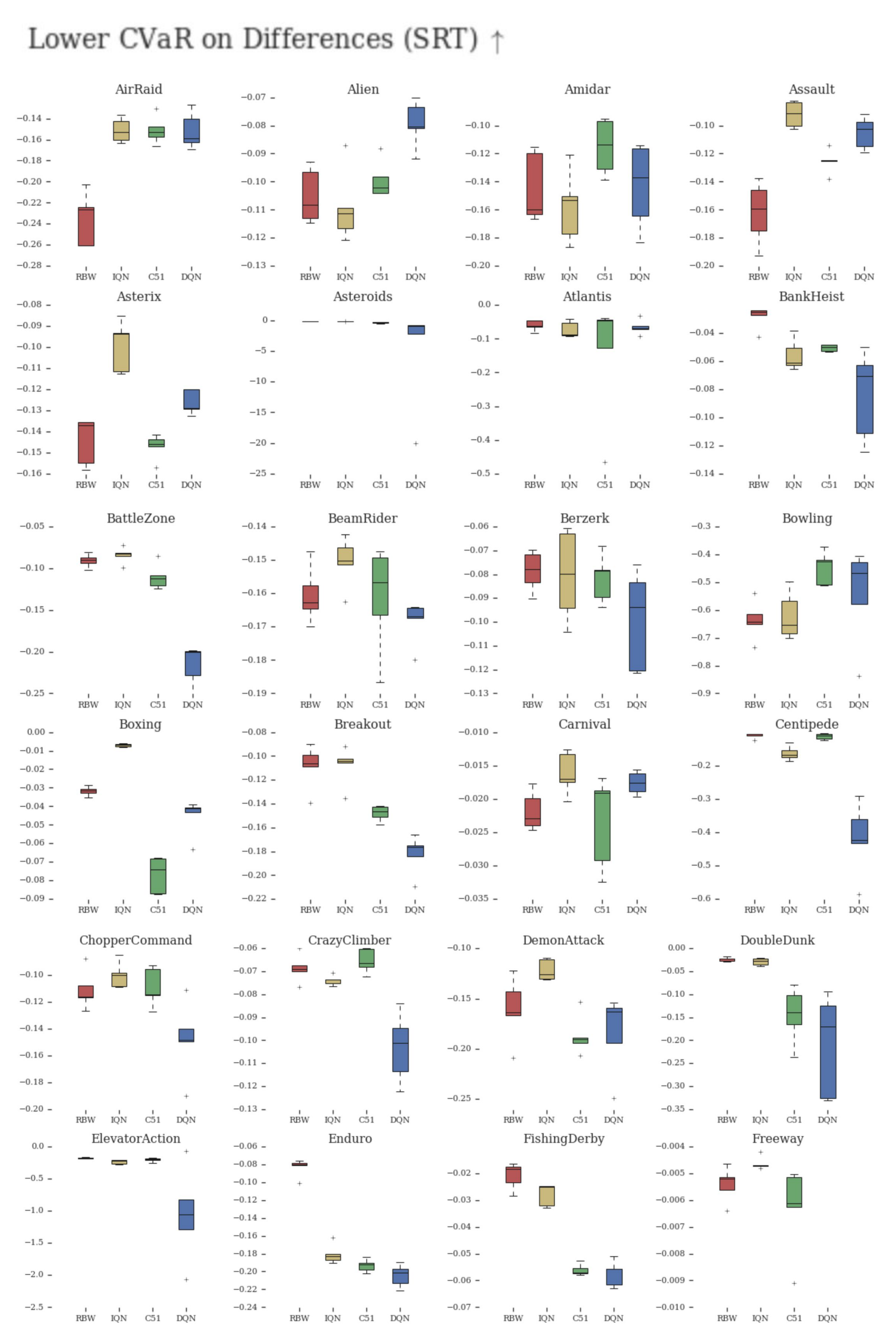}
    \caption{Short-term Risk across Time for DQN-variants tested on 60 Atari games, evaluated on a per-environment basis (page 1). Better reliability is indicated by less positive values. The x-axes indicate millions of Atari frames.}
    \label{atari_per_task_SRT1}
\end{figure}

\begin{figure}[!ht]
    \centering
    \includegraphics[width=\textwidth]{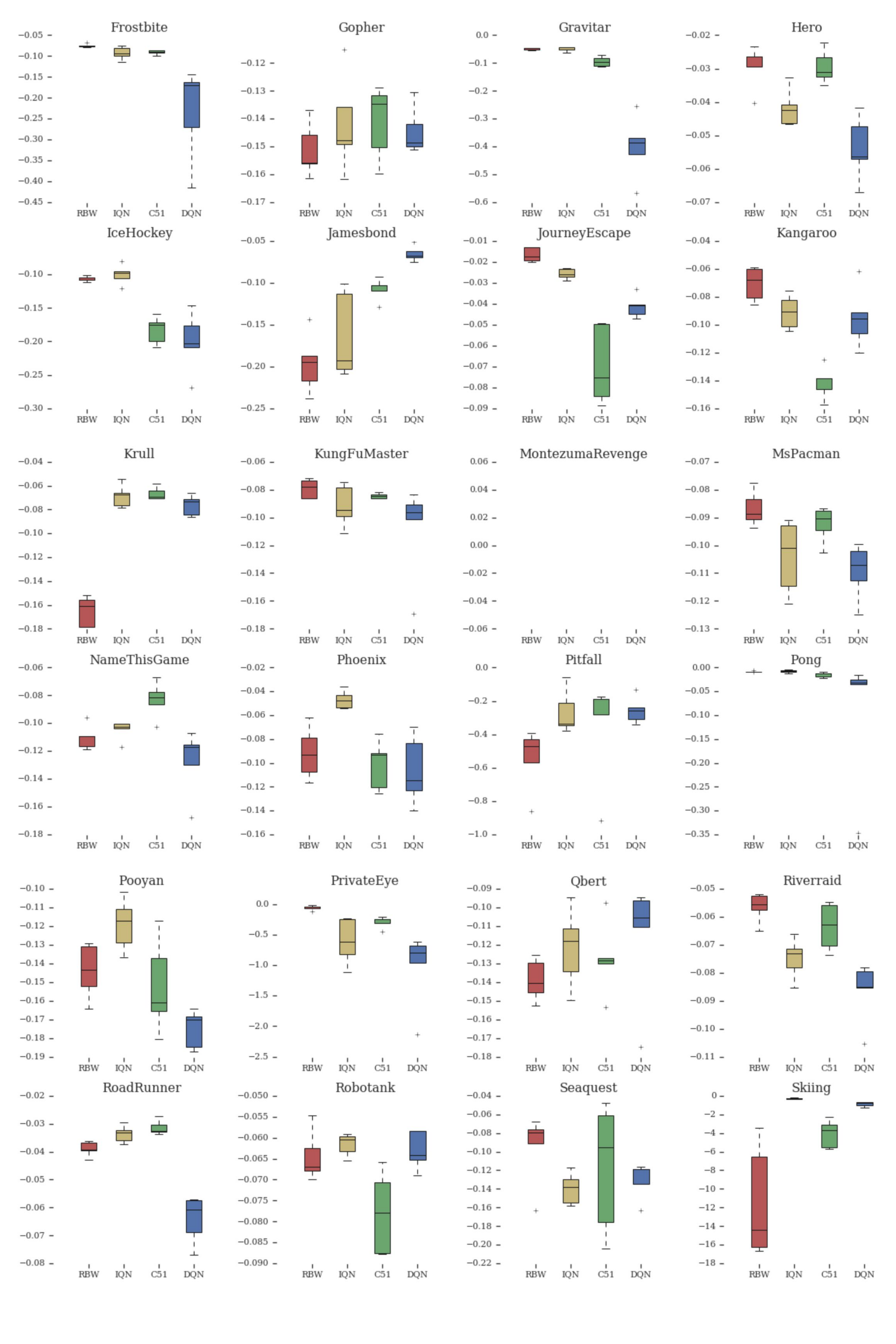}
    \caption{Short-term Risk across Time for DQN-variants tested on 60 Atari games, evaluated on a per-environment basis (page 2). Better reliability is indicated by less positive values. The x-axes indicate millions of Atari frames.}
    \label{atari_per_task_SRT2}
\end{figure}

\begin{figure}[!ht]
    \centering
    \includegraphics[width=\textwidth]{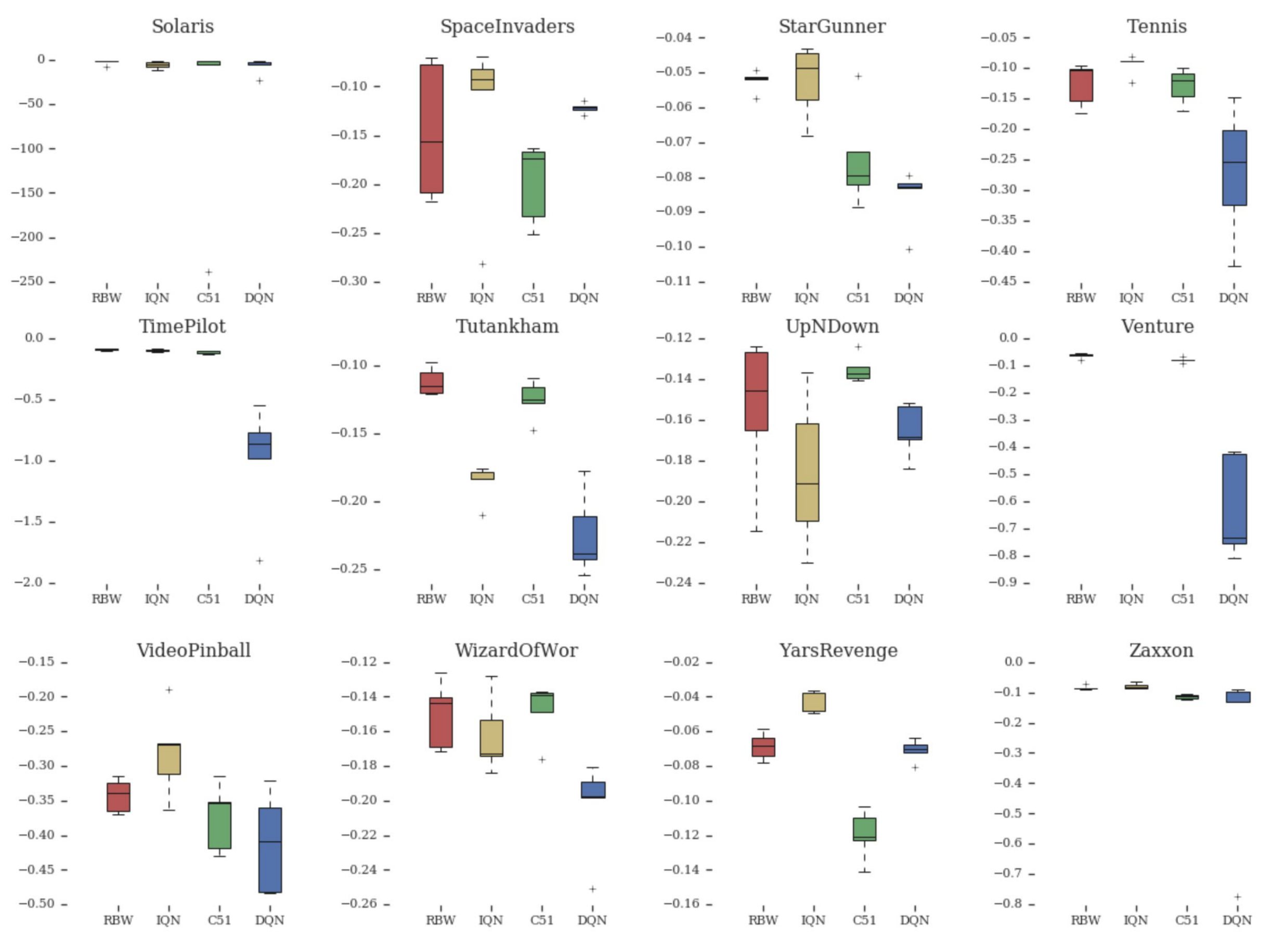}
    \caption{Short-term Risk across Time for DQN-variants tested on 60 Atari games, evaluated on a per-environment basis (page 3). Better reliability is indicated by less positive values. The x-axes indicate millions of Atari frames.}
    \label{atari_per_task_SRT3}
\end{figure}

\begin{figure}[!ht]
    \centering
    \includegraphics[width=\textwidth]{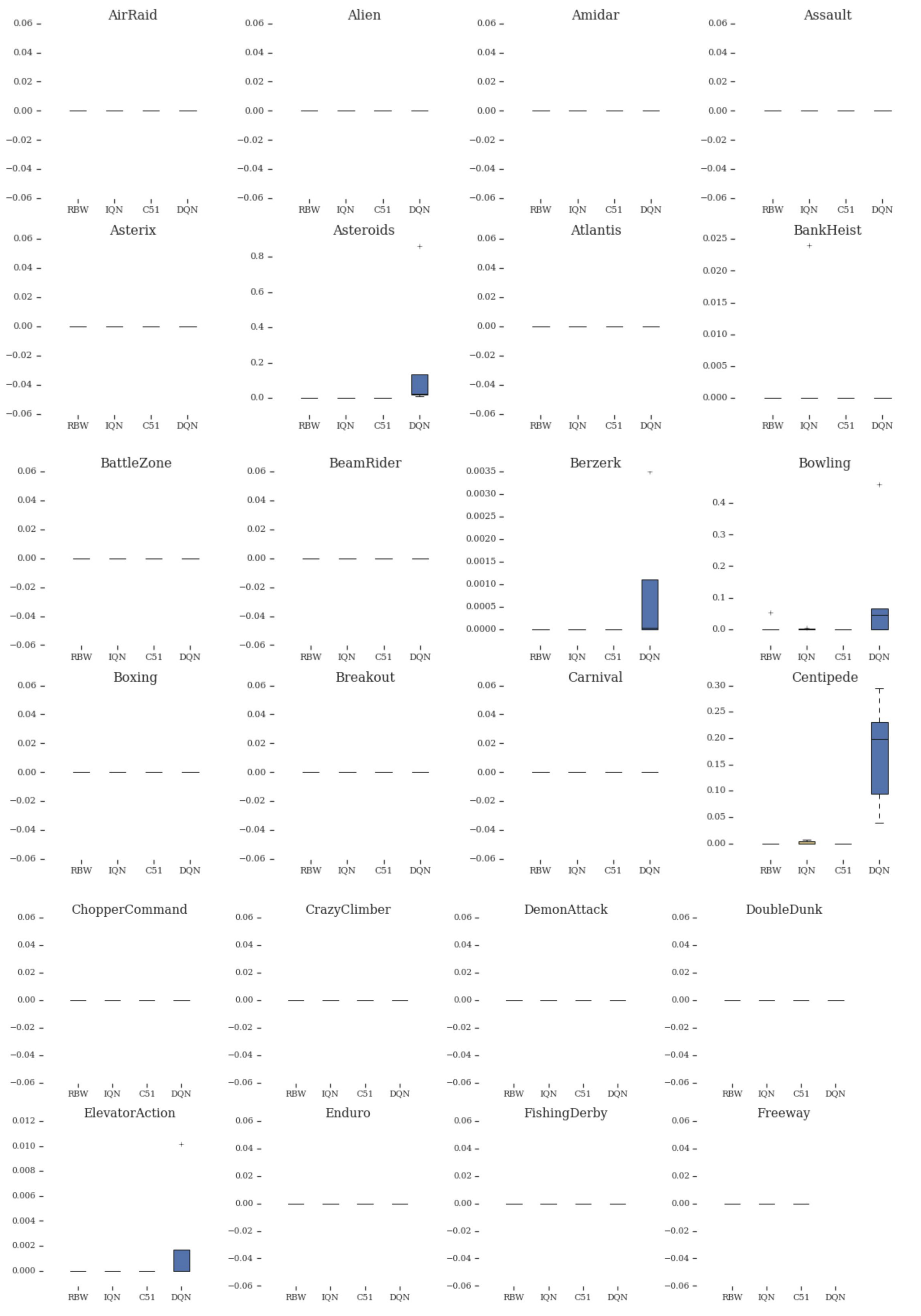}
    \caption{Long-term Risk across Time for DQN-variants tested on 60 Atari games, evaluated on a per-environment basis (page 1). Better reliability is indicated by less positive values. The x-axes indicate millions of Atari frames.}
    \label{atari_per_task_LRT1}
\end{figure}

\begin{figure}[!ht]
    \centering
    \includegraphics[width=\textwidth]{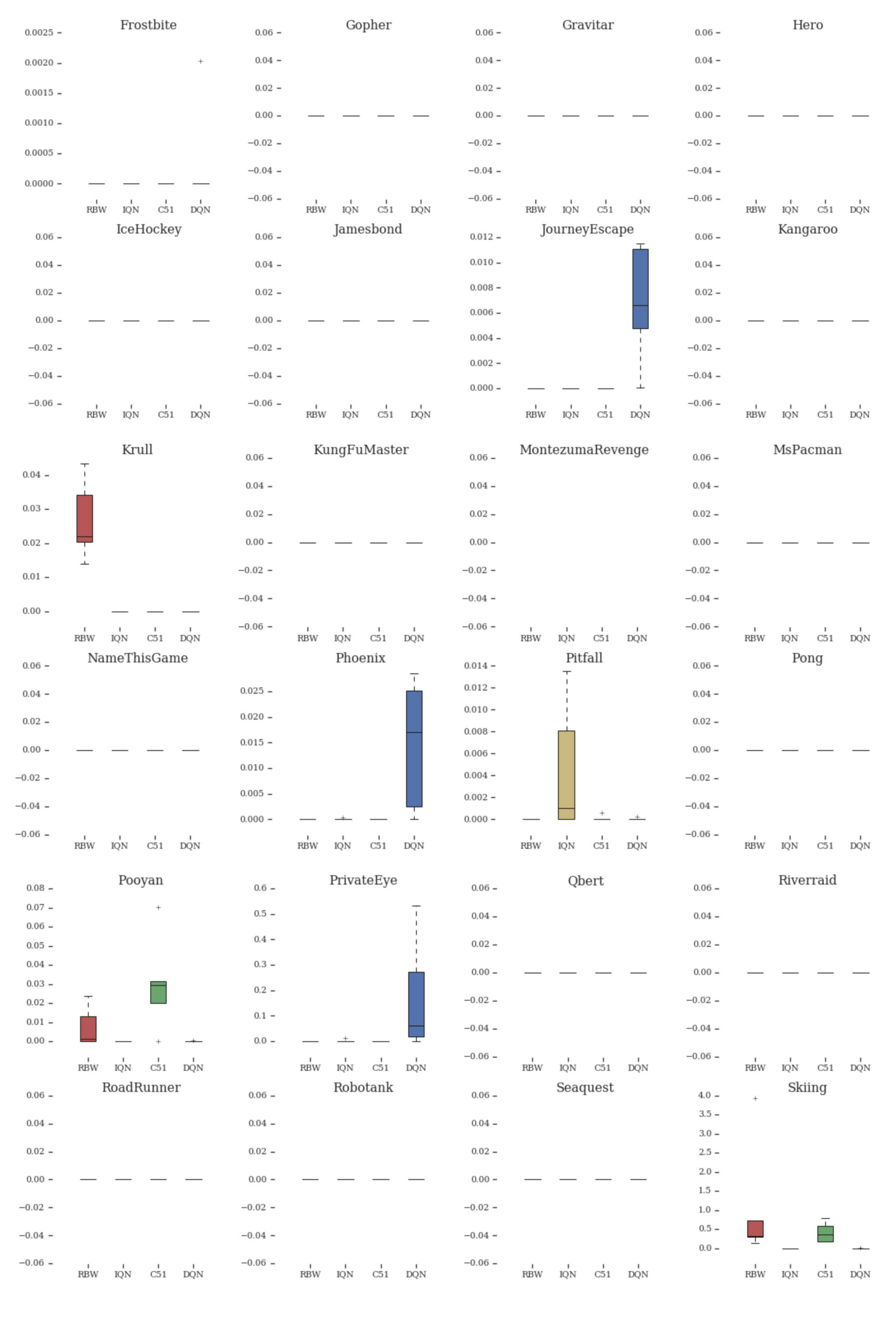}
    \caption{Long-term Risk across Time for DQN-variants tested on 60 Atari games, evaluated on a per-environment basis (page 2). Better reliability is indicated by less positive values. The x-axes indicate millions of Atari frames.}
    \label{atari_per_task_LRT2}
\end{figure}

\begin{figure}[!ht]
    \centering
    \includegraphics[width=\textwidth]{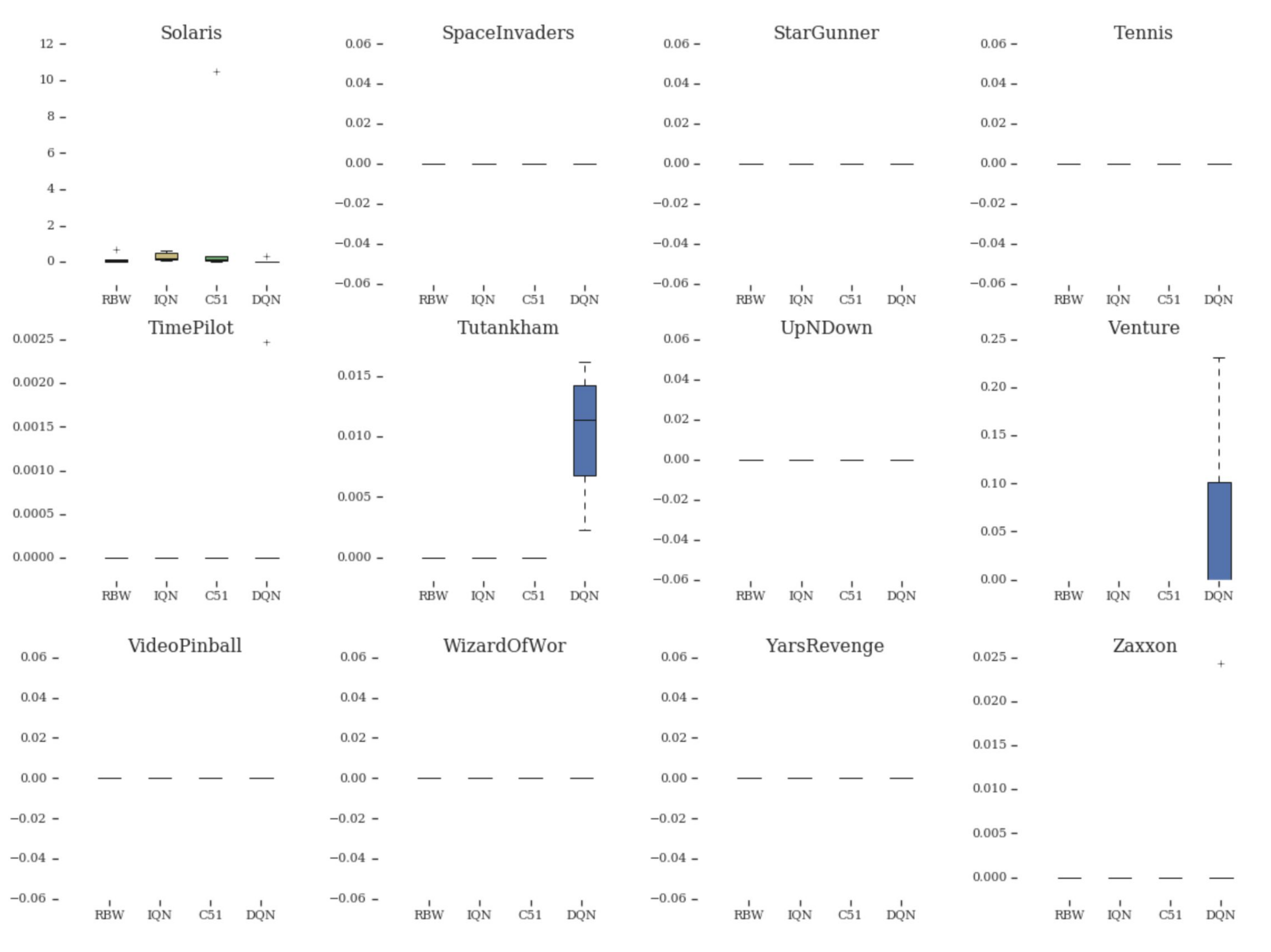}
    \caption{Long-term Risk across Time for DQN-variants tested on 60 Atari games, evaluated on a per-environment basis (page 3). Better reliability is indicated by less positive values. The x-axes indicate millions of Atari frames.}
    \label{atari_per_task_LRT3}
\end{figure}

\begin{figure}[!ht]
    \centering
    \includegraphics[width=\textwidth]{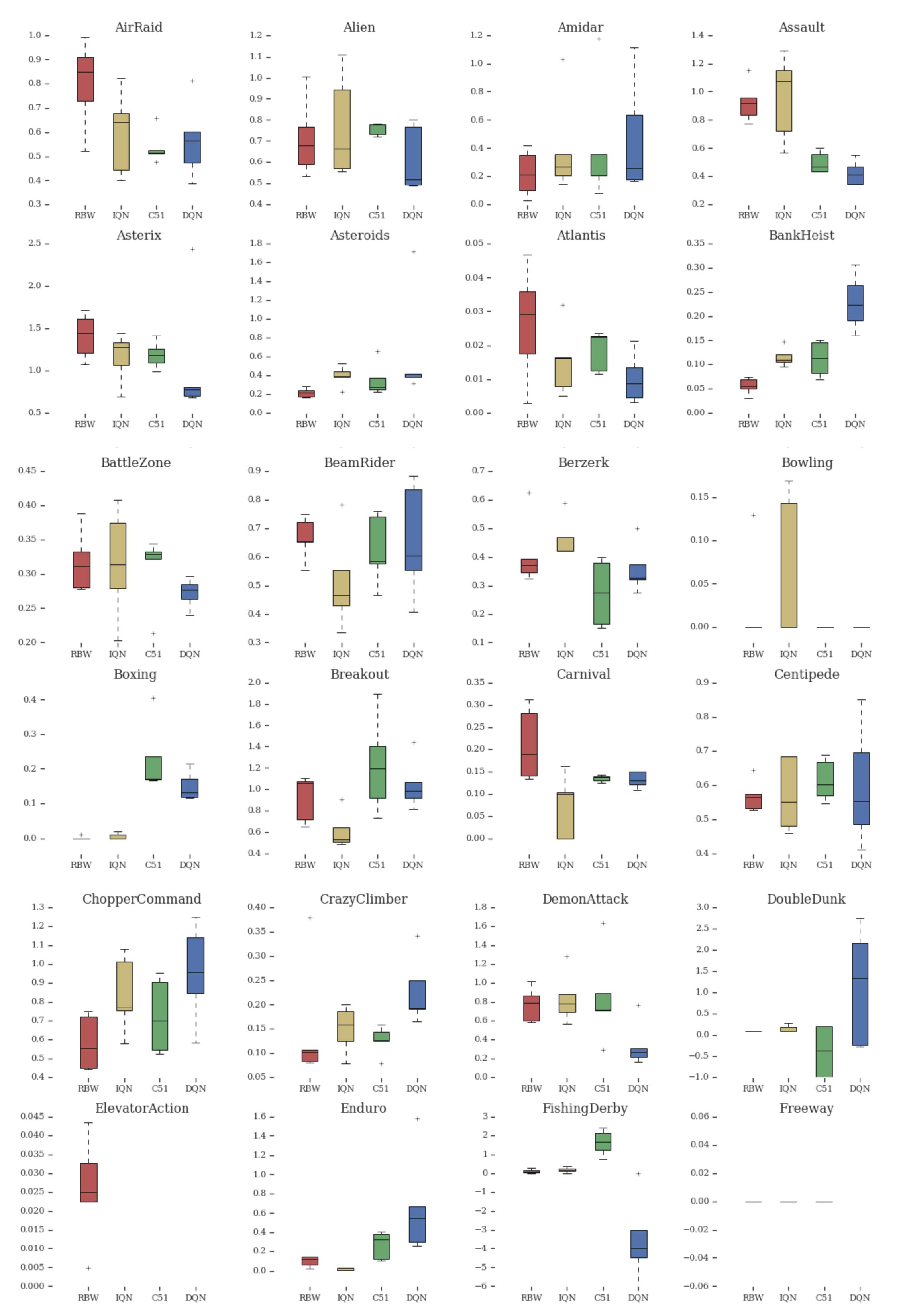}
    \caption{Dispersion across Fixed-policy Rollouts for DQN-variants tested on 60 Atari games, evaluated on a per-environment basis (page 1). Better reliability is indicated by less positive values. The x-axes indicate millions of Atari frames.}
    \label{atari_per_task_DF1}
\end{figure}

\begin{figure}[!ht]
    \centering
    \includegraphics[width=\textwidth]{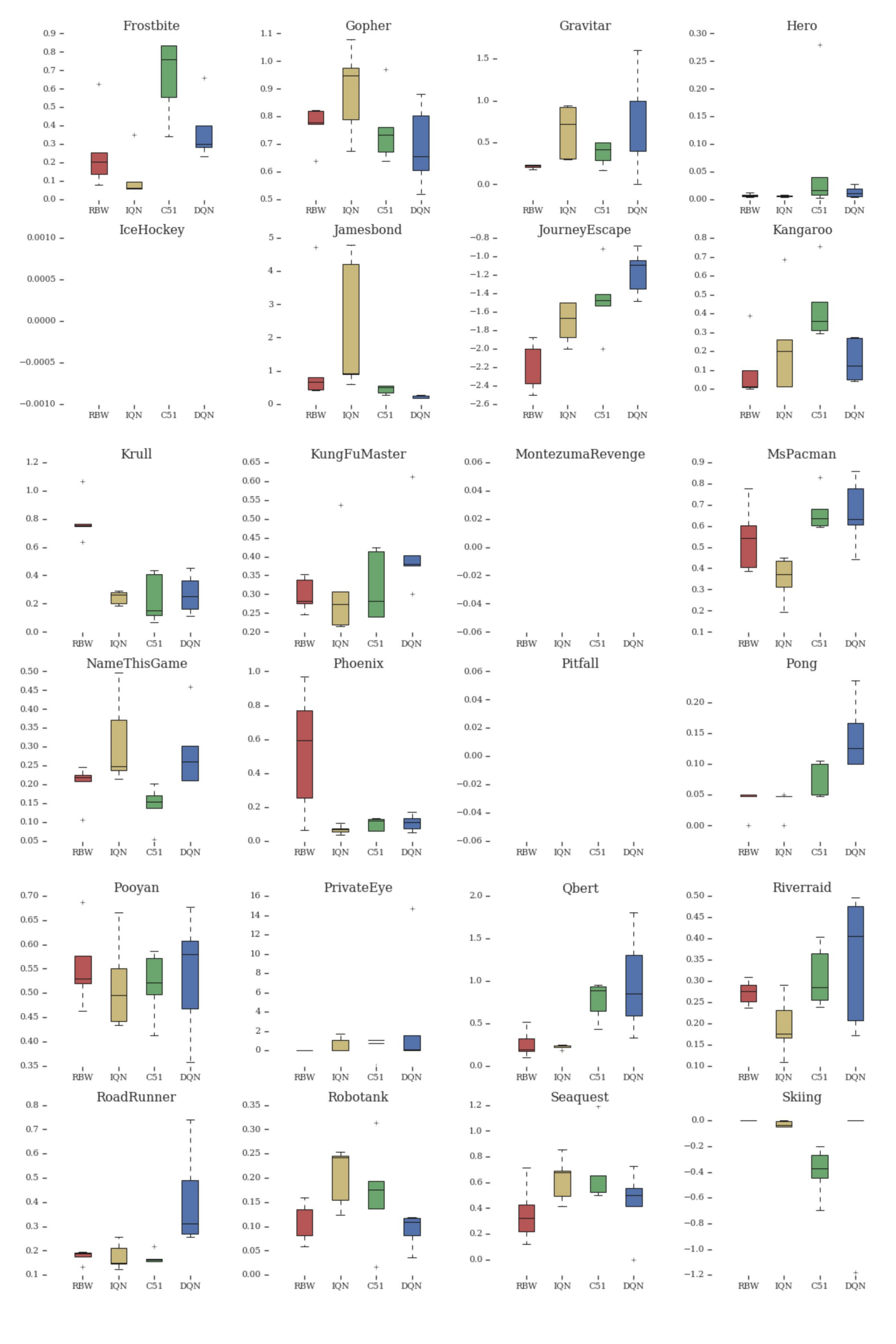}
    \caption{Dispersion across Fixed-policy Rollouts for DQN-variants tested on 60 Atari games, evaluated on a per-environment basis (page 2). Better reliability is indicated by less positive values. The x-axes indicate millions of Atari frames.}
    \label{atari_per_task_DF2}
\end{figure}

\begin{figure}[!ht]
    \centering
    \includegraphics[width=\textwidth]{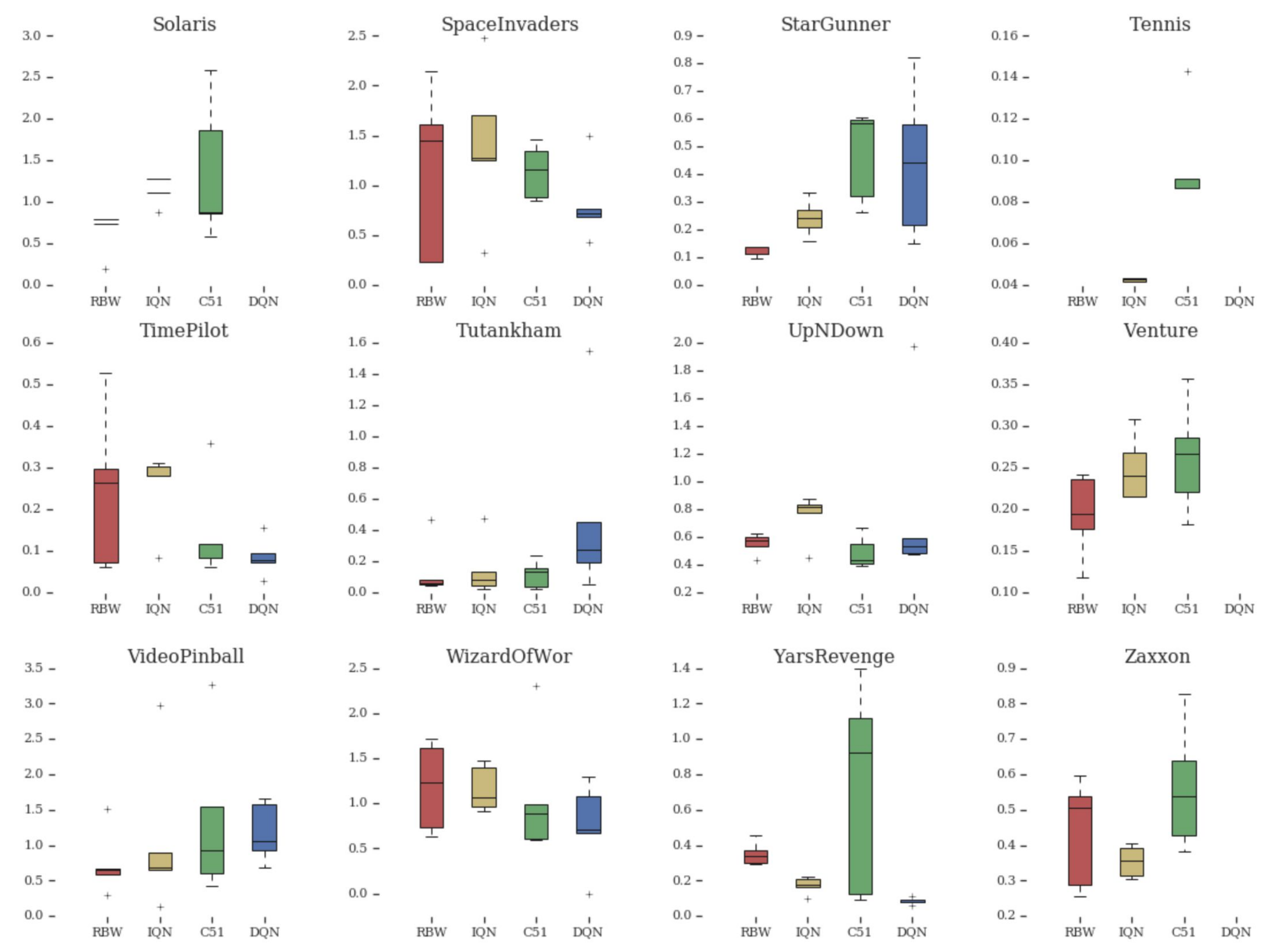}
    \caption{Dispersion across Fixed-policy Rollouts for DQN-variants tested on 60 Atari games, evaluated on a per-environment basis (page 3). Better reliability is indicated by less positive values. The x-axes indicate millions of Atari frames.}
    \label{atari_per_task_DF3}
\end{figure}

\begin{figure}[!ht]
    \centering
    \includegraphics[width=\textwidth]{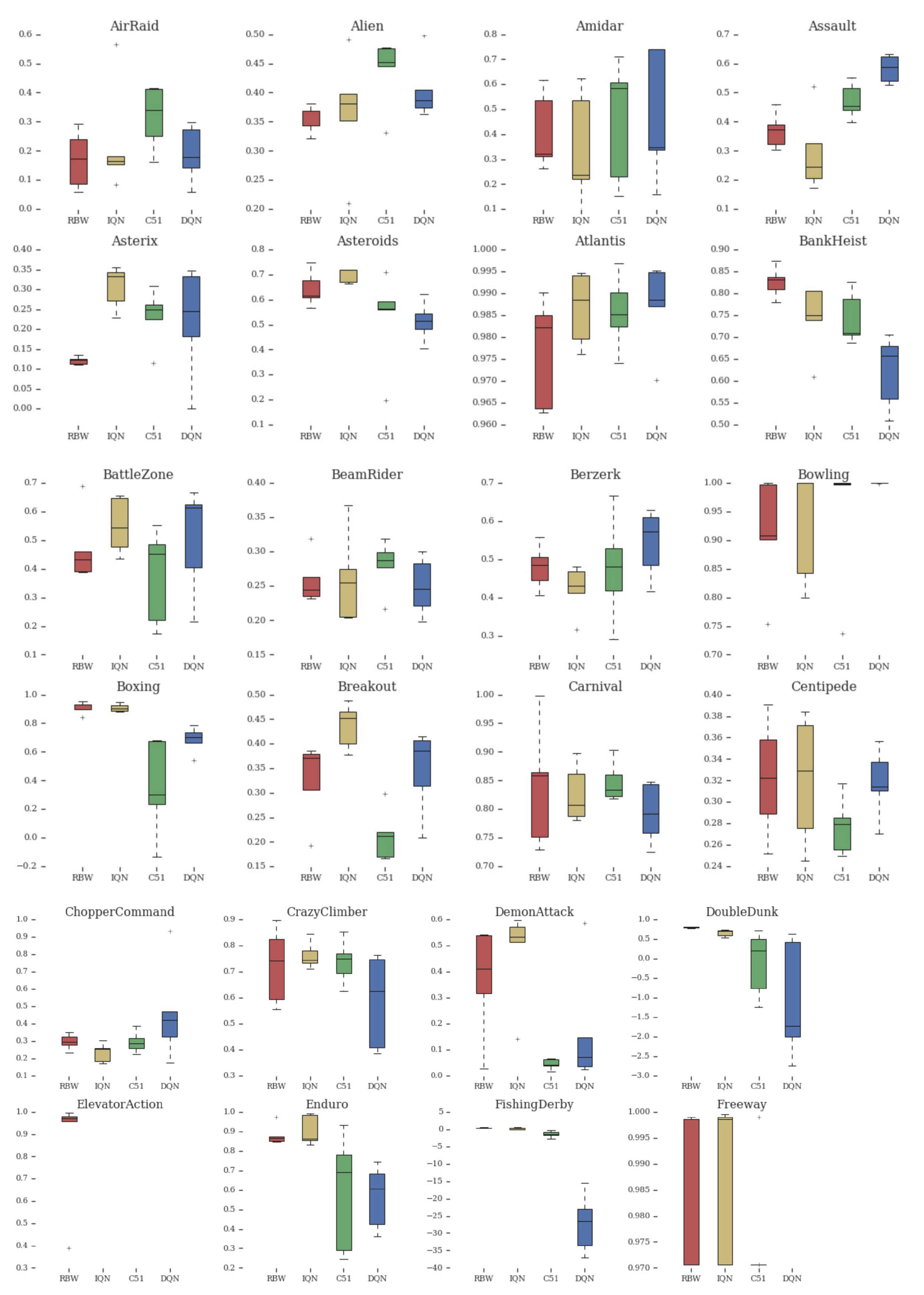}
    \caption{Risk across Fixed-policy Rollouts for DQN-variants tested on 60 Atari games, evaluated on a per-environment basis (page 1). Better reliability is indicated by more positive values. The x-axes indicate millions of Atari frames.}
    \label{atari_per_task_RF1}
\end{figure}

\begin{figure}[!ht]
    \centering
    \includegraphics[width=\textwidth]{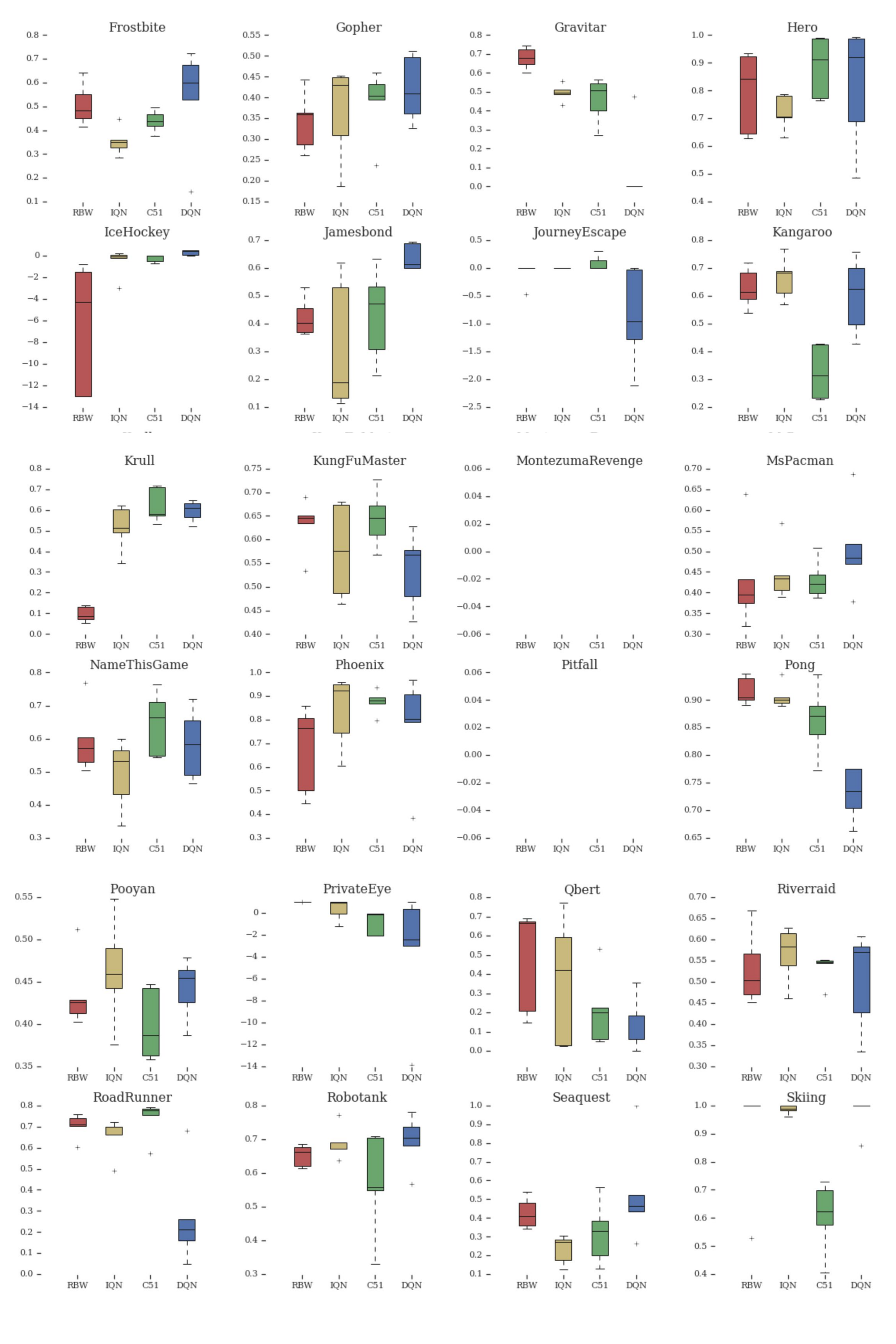}
    \caption{Risk across Fixed-policy Rollouts for DQN-variants tested on 60 Atari games, evaluated on a per-environment basis (page 2). Better reliability is indicated by more positive values. The x-axes indicate millions of Atari frames.}
    \label{atari_per_task_RF2}
\end{figure}

\begin{figure}[!ht]
    \centering
    \includegraphics[width=\textwidth]{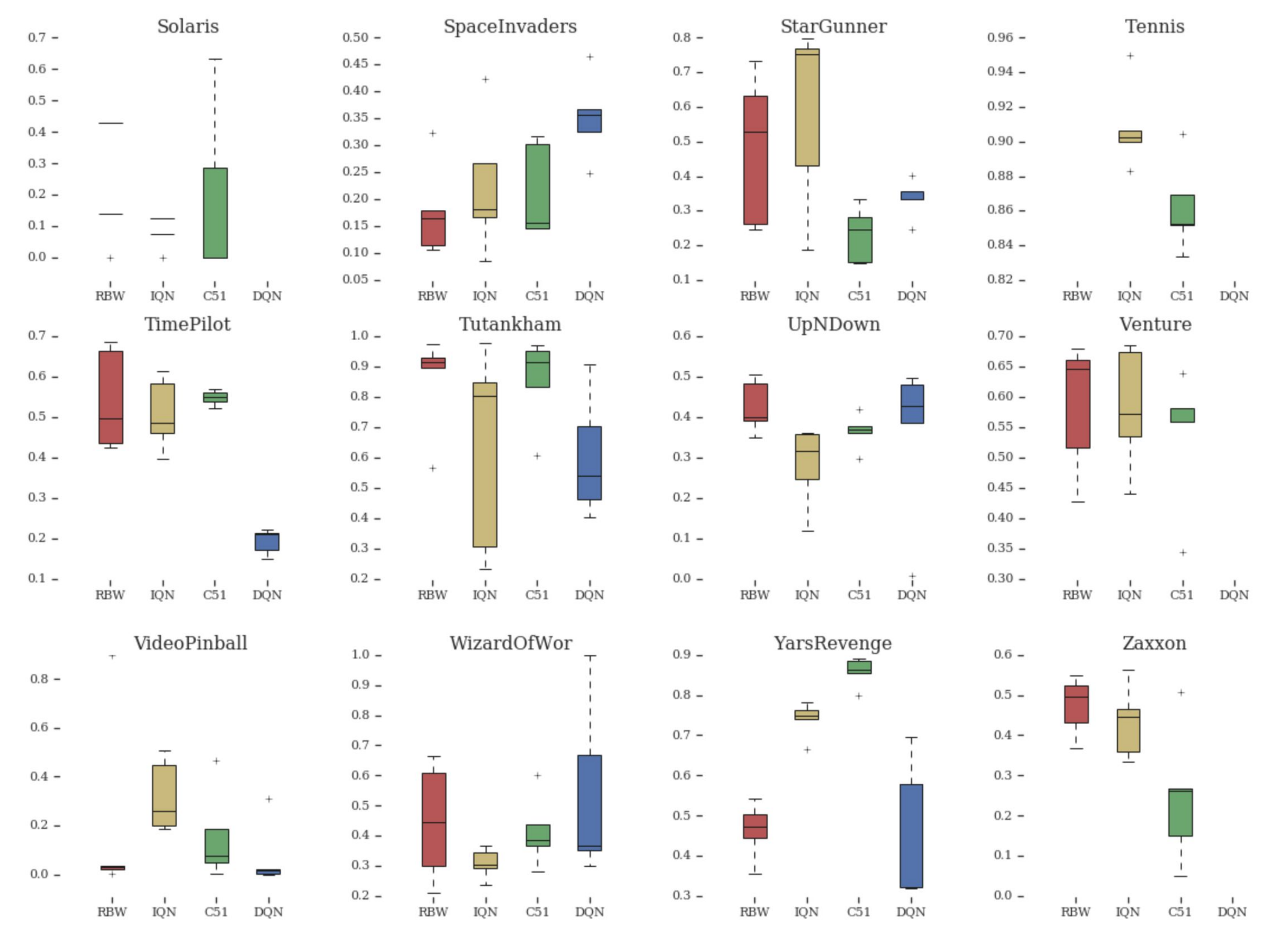}
    \caption{Risk across Fixed-policy Rollouts for DQN-variants tested on 60 Atari games, evaluated on a per-environment basis (page 3). Better reliability is indicated by more positive values. The x-axes indicate millions of Atari frames.}
    \label{atari_per_task_RF3}
\end{figure}

Metric results are shown on a per-task basis in Figs. \ref{mujoco_per_task_across_time} to \ref{mujoco_per_task_fixed_rollouts} for the OpenAI Gym MuJoCo tasks, and Figs. \ref{atari_per_task_DT1} to \ref{atari_per_task_RF3} for the Atari environments. Note that because we are no longer aggregating across tasks in this analysis, we do not need to convert the metric values to rankings.

\end{document}